 \documentclass[10pt,journal,letterpaper,compsoc]{IEEEtran}

\usepackage{algorithm,algorithmic}
\usepackage[mathscr]{eucal}
\usepackage[cmex10]{amsmath}
\usepackage{epsfig,epsf,psfrag}
\usepackage{amssymb,amsthm,amsfonts,latexsym}
\usepackage{graphicx,bm,xcolor,url}
\usepackage[caption=false]{subfig} 
\usepackage{fixltx2e}
\usepackage{array}
\usepackage{verbatim}
\usepackage{bm}
\usepackage{algorithmic, cite}
\usepackage{algorithm}
\usepackage{verbatim}
\usepackage{textcomp}
\usepackage{mathrsfs}

\catcode`~=11 \def\UrlSpecials{\do\~{\kern -.15em\lower .7ex\hbox{~}\kern .04em}} \catcode`~=13 

\allowdisplaybreaks[1]
 
\newcommand{\nn}{\nonumber}

\newcommand{\defeq}{\triangleq}
\newcommand{\cst}{\mathrm{cst}}
\newcommand{\repmat}{\mathrm{repmat}}

\newcommand{\Keff}{K_{\mathrm{eff}}}

\newcommand{\calE}{\mathcal{E}}

\newcommand{\calG}{\mathcal{G}}
\newcommand{\calH}{\mathcal{H}}
\newcommand{\calI}{\mathcal{I}}

\newcommand{\calN}{\mathcal{N}}

\newcommand{\bA}{\mathbf{A}}

\newcommand{\bB}{\mathbf{B}}

\newcommand{\bH}{\mathbf{H}}

\newcommand{\bV}{\mathbf{V}}
\newcommand{\bw}{\mathbf{w}}
\newcommand{\bW}{\mathbf{W}}
\newcommand{\bx}{\mathbf{x}}

\newcommand{\bbE}{\mathbb{E}}

\newcommand{\bbR}{\mathbb{R}}

\DeclareMathAlphabet{\mathbsf}{OT1}{cmss}{bx}{n}
\DeclareMathAlphabet{\mathssf}{OT1}{cmss}{m}{sl}

\newcommand{\rvN}{\mathsf{N}}

\newcommand{\rvR}{\mathsf{R}}

\newcommand{\rvS}{\mathsf{S}}

\DeclareSymbolFont{bsfletters}{OT1}{cmss}{bx}{n}  
\DeclareSymbolFont{ssfletters}{OT1}{cmss}{m}{n}
\DeclareMathSymbol{\bsfGamma}{0}{bsfletters}{'000}
\DeclareMathSymbol{\ssfGamma}{0}{ssfletters}{'000}
\DeclareMathSymbol{\bsfDelta}{0}{bsfletters}{'001}
\DeclareMathSymbol{\ssfDelta}{0}{ssfletters}{'001}
\DeclareMathSymbol{\bsfTheta}{0}{bsfletters}{'002}
\DeclareMathSymbol{\ssfTheta}{0}{ssfletters}{'002}
\DeclareMathSymbol{\bsfLambda}{0}{bsfletters}{'003}
\DeclareMathSymbol{\ssfLambda}{0}{ssfletters}{'003}
\DeclareMathSymbol{\bsfXi}{0}{bsfletters}{'004}
\DeclareMathSymbol{\ssfXi}{0}{ssfletters}{'004}
\DeclareMathSymbol{\bsfPi}{0}{bsfletters}{'005}
\DeclareMathSymbol{\ssfPi}{0}{ssfletters}{'005}
\DeclareMathSymbol{\bsfSigma}{0}{bsfletters}{'006}
\DeclareMathSymbol{\ssfSigma}{0}{ssfletters}{'006}
\DeclareMathSymbol{\bsfUpsilon}{0}{bsfletters}{'007}
\DeclareMathSymbol{\ssfUpsilon}{0}{ssfletters}{'007}
\DeclareMathSymbol{\bsfPhi}{0}{bsfletters}{'010}
\DeclareMathSymbol{\ssfPhi}{0}{ssfletters}{'010}
\DeclareMathSymbol{\bsfPsi}{0}{bsfletters}{'011}
\DeclareMathSymbol{\ssfPsi}{0}{ssfletters}{'011}
\DeclareMathSymbol{\bsfOmega}{0}{bsfletters}{'012}
\DeclareMathSymbol{\ssfOmega}{0}{ssfletters}{'012}

\newcommand{\hatb}{\hat{b}}

\newcommand{\tilF}{J}

\newcommand{\tilh}{\tilde{h}}

\newcommand{\hatv}{\hat{v}}

\newcommand{\blambda}{\bm{\lambda}}

\newcommand{\tlambda}{\tilde{\lambda}}

\newcommand{\hmu}{\hat{\mu}}

\def\fndot{\, \cdot \,}

\DeclareMathOperator*{\argmin}{arg\,min}

\DeclareMathOperator{\minimize}{minimize}

\DeclareMathOperator{\st}{subject\,\,to}

\DeclareMathOperator{\diag}{diag}

\DeclareMathOperator{\var}{var}

\newtheorem{theorem}{Theorem} 
\newtheorem{lemma}[theorem]{Lemma}

\usepackage{epstopdf}
\usepackage{url}
\usepackage{cite}
\usepackage{ulem,color}
\def\bal#1{\begin{align}#1\end{align}}

\def\argmin{\mathop{\mathrm{arg\,min}}}
\newcommand{\ve}[1]{ {\mathbf{#1}} }

\newcommand{\SNR}{\rvS\rvN\rvR}

\title{Automatic Relevance Determination in Nonnegative Matrix Factorization with the $\beta$-Divergence}
\author{Vincent Y.~F.~Tan, \IEEEmembership{Member, IEEE} $\,$ and $\,$ C\'{e}dric F\'{e}votte, \IEEEmembership{Member, IEEE}  \thanks{The work of V.~Y.~F.~Tan is supported by A*STAR, Singapore. The work of C.~ F\'{e}votte is supported by project ANR-09-JCJC-0073-01 TANGERINE (Theory and applications of nonnegative matrix factorization).} \thanks{V.~Y.~F.~Tan is with the  Institute for Infocomm Research, A*STAR, Singapore and the Department Electrical and Comptuer Engineering, National University of Singapore (emails: tanyfv@i2r.a-star.edu.sg, eletyfv@nus.edu.sg).  C.~F\'{e}votte  is with CNRS LTCI, T\'el\'ecom ParisTech, 75014 Paris, France (e-mail: fevotte@telecom-paristech.fr). } }

\IEEEcompsoctitleabstractindextext{
\begin{abstract}
This paper addresses the  estimation of the latent dimensionality in nonnegative matrix factorization (NMF) with the $\beta$-divergence.  The $\beta$-divergence is a family of cost functions that includes the squared Euclidean distance, Kullback-Leibler and Itakura-Saito divergences as special cases. Learning the model order is important as it is necessary to strike the right balance between data fidelity and overfitting. We propose a Bayesian model based on {\em automatic relevance determination} in which the columns of the dictionary matrix and the rows of the activation matrix are tied together through a common scale parameter in their prior. A family of majorization-minimization algorithms is proposed for  maximum a posteriori (MAP) estimation. A subset of scale parameters is driven to a small lower bound in the course of inference, with the effect of pruning the corresponding spurious components. We demonstrate the efficacy and robustness of our algorithms by performing extensive experiments on synthetic  data, the \texttt{swimmer} dataset, a music decomposition example and a stock price prediction task.

\end{abstract} 
 }

\begin{document}
\maketitle

\section{Introduction} \label{sec:intro}

Given a  data matrix $\ve{V}$ of dimensions ${F \times N}$ with nonnegative entries, nonnegative matrix factorization (NMF) consists in finding a  low-rank factorization
\begin{equation} \label{eqn:facto}
\bV \approx \hat{\bV}\defeq\bW \bH
\end{equation}
where $\bW$ and $\bH$ are nonnegative matrices of dimensions $F \times K$ and $K \times N$, respectively. The common dimension $K$ is usually chosen such that $F\,K + K\,N \ll F\,N$, hence the overall number of parameters to describe the data (i.e., data dimension) is reduced.  Early references on NMF include the work of Paatero and Tapper~\cite{paa94} and a seminal contribution by Lee and  Seung  \cite{lee99}. Since then, NMF has become  a widely-used technique   for non-subtractive, parts-based representation of nonnegative data. There are numerous applications of NMF in diverse fields, such as audio signal processing~\cite{neco09}, image classification~\cite{Guillamet},   analysis of financial data~\cite{dra07},  and  bioinformatics~\cite{Gao05}. The factorization~\eqref{eqn:facto} is usually sought after through the minimization problem 
\begin{equation} \label{eqn:mini}
\mathop{\minimize}_{\bW,\bH}  \ D(\bV | \bW \bH) \,\,  \text{subject to} \,\, \bW \ge 0, \bH \ge 0
\end{equation}
where  $\bA \ge 0$ means that  all entries of the matrix $\bA$ are nonnegative  (and not positive semidefiniteness). The  function $D(\bV | \bW \bH)$ is a separable measure of fit, i.e., 
\begin{equation} \label{eqn:defcost}
D(\bV | \bW \bH) = \sum_{f=1}^F \sum_{n=1}^N d( [\bV]_{fn} \, | \, [\bW \bH]_{fn})
\end{equation}
where $d(x | y)$ is a scalar cost function  of $y \in \mathbb{R}_+$ given $x \in \mathbb{R}_+$; and it equals zero when $x = y$.     In this paper, we will consider the $d(x|y)$ to be the $\beta$-divergence, a family of cost functions parametrized by a single scalar $\beta\in\bbR$. The squared Euclidean (EUC) distance, the generalized Kullback-Leibler (KL) divergence and the Itakura-Saito (IS) divergence are special cases of the $\beta$-divergence. NMF with the $\beta$-divergence (or, in short, $\beta$-NMF) was first considered by Cichocki et al.\ in~\cite{cic06a}, and more detailed treatments have been proposed in~\cite{naka10, betanmf, cic11}. 

\subsection{Main Contributions}
In most applications, it is crucial that the ``right'' model order $K$ is  selected. If $K$ is too small, the data does not fit the model well. Conversely, if $K$ too large, overfitting   occurs. We   seek  to find an elegant solution for this dichotomy between data fidelity and  overfitting.  Traditional model selection techniques such as the Bayesian information criterion (BIC)~\cite{Schwarz} are not applicable in our setting as the number of parameters  is $FK+KN$ and this  scales linearly with the number of data points $N$, whereas BIC assumes that the number of parameters stays constant as the number of data points increases. 

To ameliorate this problem, we   propose a Bayesian model for $\beta$-NMF based on automatic relevance determination (ARD) \cite{mac95}, and in particular, we are  inspired by Bayesian PCA (Principal Component Analysis)~\cite{bis99}. We  derive {\em computationally efficient}  algorithms with {\em monotonicity} guarantees to estimate the  model order $K$ and to  estimate the basis $\bW$ and the activation coefficients $\bH$. The proposed algorithms are based on  the use of auxiliary functions (local majorizations of   the objective function). The optimization of these  auxiliary functions leads directly to majorization-minimization (MM) algorithms, resulting in  efficient multiplicative updates. The monotonicity of the objective function  can be proven by leveraging on techniques in~\cite{betanmf}.  We show via   simulations in Section~\ref{sec:exps} on synthetic data and real datasets (such as a music decomposition example) that the proposed algorithms   recover the correct model order and produce better decompositions. We also describe a procedure based on the \emph{method of moments} for adaptive and data-dependent selection of some of the hyperparameters.

\subsection{Prior Work}
To the best of our knowledge, there is  fairly limited literature on   model   order selection in NMF. References~\cite{cem09} and~\cite{schmidt09} describe Markov chain Monte Carlo (MCMC) strategies for evaluation of the model evidence in EUC-NMF or KL-NMF. The evidence is calculated for each   candidate value  of $K$, and the model with highest evidence is selected. References~\cite{zhong09,schmidt10} describe reversible jump MCMC approaches that allow to sample the model order $K$, along with any other parameter. These sampling-based methods   are  computationally   intensive. Another class of methods, given in references~\cite{mor09,mor09b,yang10,hoffman}, is closer to the principles that underlie this work; in these works the number of components $K$ is set to a large value and irrelevant components in $\bW$ and $\bH$ are driven to zero during  inference. A detailed but qualitative comparison between our work and these   methods is given in Section~\ref{sec:connect}. In Section~\ref{sec:exps}, we compare the empirical performance of our methods to~\cite{mor09} and~\cite{hoffman}.

This paper is a significant extension of the authors' conference publication in~\cite{spars09}. Firstly, the cost function in~\cite{spars09} was restricted to be the  KL-divergence. In this paper, we consider a continuum of costs parameterized by $\beta$, underlying  different statistical  noise models. We show that this flexibility in the cost function  allows for better quality of factorization and model   selection on various classes of real-world signals such as audio and images. Secondly, the algorithms described herein are such that the cost function monotonically decreases to a local minimum whereas the algorithm in~\cite{spars09} is   heuristic. Convergence is guaranteed by the MM framework.

\subsection{Paper organization}
In Section~\ref{sec:prelim}, we state our notation and introduce $\beta$-NMF and the MM technique. In Section~\ref{sec:ardnmf}, we present our Bayesian model for $\beta$-NMF. Section~\ref{sec:algos} details $\ell_1$- and $\ell_2$-ARD for model selection in $\beta$-NMF. We then compare the proposed algorithms to other related works in Section~\ref{sec:connect}.  In Section~\ref{sec:exps}, we present extensive numerical results to demonstrate the efficacy and robustness of  $\ell_1$- and $\ell_2$-ARD. We conclude the discussion in Section~\ref{sec:concl}. 

\section{Preliminaries} \label{sec:prelim}

\subsection{Notations} \label{sec:notations}
We denote by $\bV$, $\bW$ and $\bH$, the data, dictionary and activation matrices, respectively. These nonnegative matrices are  of dimensions $F \times N$, $F \times K$ and $K \times N$, respectively. The entries of these matrices are denoted by $v_{fn}$, $w_{fk}$ and $h_{kn}$ respectively. The $k^{\mathrm{th}}$ \emph{column} of $\bW$ is denoted by $\bw_k \in \bbR_+^{F}$, and $\underline{h}_k \in \bbR_+^{N}$ denotes the $k^{\mathrm{th}}$ \emph{row} of $\bH$.  Thus, $\bW=[\bw_1,\ldots, \bw_K]$ and $\bH=[\underline{h}_1^T,\ldots, \underline{h}_K^T]^T$.

\subsection{NMF with the $\beta$-divergence} \label{sec:betanmf}

This paper considers NMF based on the $\beta$-divergence, which we now review. The $\beta$-divergence was originally introduced for $\beta\ge 1$ in \cite{basu98, egu01} and later generalized to $\beta\in\bbR$ in \cite{cic06a}, which is the definition we use here: 
\begin{equation} \label{eqn:beta}
d_{\beta}(x | y) \defeq
\left\lbrace
\begin{array}{cl}
    \frac{x^{\beta}}{{\beta\,(\beta-1)}} + \frac{y^{\beta}}{\beta} -  \frac{ x\, y^{\beta-1}}{\beta-1}  &  \beta \in \mathbb{R} \backslash \{0,1\}  \\
x\, \log \frac{x}{y} - x + y & \beta = 1 \\
\frac{x}{y} - \log \frac{x}{y} - 1 & \beta = 0
\end{array}
\right.
\end{equation}
The limiting cases $\beta = 0$ and $\beta = 1$ correspond to the IS and KL-divergences, respectively. Another case of note is $\beta=2$ which corresponds to the squared Euclidean distance, i.e., $d_{\beta=2}(x|y)=(x-y)^2/2$. The parameter $\beta$ essentially controls the assumed statistics of the observation noise and can either be fixed  or learnt from training data   by cross-validation. Under certain assumptions, the $\beta$-divergence can be mapped to a log-likelihood function for the Tweedie distribution \cite{twee84}, parametrized with respect to its mean.   In particular, the values $\beta =0,1,2$ underlie the multiplicative Gamma observation noise, Poisson noise  and  Gaussian additive  observation noise respectively. We describe this property in greater detail in Section~\ref{sec:likel}. The $\beta$-divergence offers a continuum of noise statistics that interpolates between these three specific cases. In the following, we use the notation $D_\beta(\bV|\bW \bH)$ to denote the separable cost function in \eqref{eqn:defcost} with the scalar cost  $d=d_{\beta}$ in \eqref{eqn:beta}. 

\begin{table}
{
\renewcommand{\arraystretch}{1.5}
\begin{tabular}{|l|l|} \hline
Auxiliary function $G(\bH | \tilde{\bH})$ & $\beta$  \\\hline
$\sum_{kn} q_{kn} h_{kn} - \frac{1}{\beta-1} p_{kn} \tilde{h}_{kn} \left( \frac{h_{kn}}{\tilde{h}_{kn}} \right)^{\beta-1} + \cst$ & $\beta<1$ \\\hline
$\sum_{kn} q_{kn} h_{kn}-p_{kn}\tilh_{kn}\log \left(\frac{h_{kn}}{\tilde{h}_{kn}} \right) \! + \!\cst$ & $\beta=1$ \\\hline
$\sum_{kn} \frac{1}{\beta} q_{kn} \tilde{h}_{kn} \left( \frac{h_{kn}}{\tilde{h}_{kn}} \right)^{\beta}   \!\! -\! \frac{1}{\beta-1} p_{kn} \tilde{h}_{kn} \left( \frac{h_{kn}}{\tilde{h}_{kn}} \right)^{\beta-1} \!+\! \cst$ & $\beta \in  (1,2]$ \\\hline
$\sum_{kn} \frac{1}{\beta} q_{kn} \tilde{h}_{kn} \left( \frac{h_{kn}}{\tilde{h}_{kn}} \right)^{\beta}    - p_{kn} \tilde{h}_{kn}  \!+\! \cst$ & $\beta>2$ \\\hline
\end{tabular} }
\caption{The form of the auxiliary function for various $\beta$'s \cite{betanmf}. } 
\label{tab:aux}
\end{table} 

\subsection{Majorization-minimization (MM) for $\beta$-NMF} \label{sec:mm}
We briefly recall some results   in~\cite{betanmf} on standard $\beta$-NMF. In particular, we describe how an MM  algorithm \cite{hunt04} that recovers a stationary point of \eqref{eqn:defcost} can be derived. The algorithm  updates $\bH$ given $\bW$, and $\bW$ given $\bH$, and these two steps are essentially the same by the symmetry of $\bW$ and $\bH$ by transposition ($\bV\approx \bW \bH$ is equivalent to $\bV^T \approx \bH^T \bW^T$). Let us thus focus on  the optimization of $\bH$ given $\bW$. The MM framework involves building a (nonnegative) \emph{auxiliary function} $G(\bH|\tilde{\bH})$ that majorizes the objective $C(\bH) = D_\beta(\bV|\bW\bH)$ everywhere, i.e., 
\begin{equation}
G(\bH|\tilde{\bH}) \ge C(\bH), \label{eqn:major}
\end{equation}
for all pairs of nonnegative matrices $\bH,\tilde{\bH}\in\bbR_+^{K\times N}$. The auxiliary function also matches the cost function whenever its arguments are the same, i.e., for all $\tilde{\bH}$,  
\begin{equation}
G(\tilde{\bH}|\tilde{\bH})= C(\tilde{\bH}) . \label{eqn:eqaulity}
\end{equation}
If such an auxiliary function exists and the optimization of  $G(\bH|\tilde{\bH})$ over $\bH$ for fixed $\tilde{\bH}$ is   simple, the optimization of $C(\bH)$ may be replaced by the simpler optimization of $G(\bH|\tilde{\bH})$ over $\bH$. Indeed, any iterate $\bH^{(i+1)}$ such that $G(\bH^{(i+1)}| \bH^{(i)}) \le G(\bH^{(i)}| \bH^{(i)})$ reduces the   cost since
\begin{equation}
C(\bH^{(i+1)}) \le G(\bH^{(i+1)} | \bH^{(i)}) \le G(\bH^{(i)} | \bH^{(i)}) = C(\bH^{(i)}). \label{eqn:inequalities}
\end{equation}
The first inequality follows from \eqref{eqn:major} and the second from the optimality of $\bH^{(i+1)}$. The MM update thus consists in 
\begin{equation} \label{eqn:max}
\bH^{(i+1)} = \argmin_{\bH \ge 0} \ G(\bH | \bH^{(i)}) .
\end{equation}
Note that if $\bH^{(i+1)}=\bH^{(i)}$, a local minimum is attained since the inequalities in \eqref{eqn:inequalities} are equalities.  The key of the MM approach is thus to build an auxiliary function $G$ which reasonably approximates the original objective at the current iterate $\tilde{\bH}$, and such that the function is easy to minimize (over the first variable $\bH$). In our setting, the objective function $C(\bH)$ can be decomposed into the sum of a convex term and a concave term. As such, the construction proposed in \cite{betanmf} and \cite{naka10} consists in majorizing the convex and concave terms separately, using Jensen's inequality and a first-order Taylor approximation, respectively. Denoting $\tilde{v}_{fn} \defeq [\bW \tilde{\bH}]_{fn}$ and 
\begin{align}
   p_{kn}  \defeq \sum_f w_{fk} v_{fn} \tilde{v}_{fn}^{\beta-2} ,\qquad  q_{kn} \defeq \sum_f w_{fk}  \tilde{v}_{fn}^{\beta-1}\label{eqn:qkn}
\end{align}
the resulting auxiliary function can be expressed as in Table~\ref{tab:aux},  
where $\cst$ denote  constant terms that do not depend on  $\bH$. In the sequel, the use of the tilde over a parameter will generally denote its  {\em previous}  iterate.  Minimization of $G(\bH | \tilde{\bH})$ with respect to (w.r.t) $\bH$  thus leads to the following simple update
\begin{equation} \label{eqn:update_hkn}
h_{kn} = \tilde{h}_{kn} \left( \frac{p_{kn}}{q_{kn}} \right)^{\gamma(\beta)}
\end{equation}
where the exponent $\gamma(\beta)$ is defined as 
\begin{equation} \label{eqn:defgamma}
\gamma(\beta) \defeq
\left\{
\begin{array}{ll}
{1}/(2-\beta), & \beta <1 \\
1, &  1 \le \beta \le 2 \\
1/(\beta-1), & \beta > 2
\end{array}
\right. 
\end{equation}

\section{The Model for Automatic relevance determination in $\beta$-NMF} \label{sec:ardnmf}
In this section, we describe our probabilistic  model for NMF. The model involves tying the  $k^{\mathrm{th}}$ column of $\bW$ to the $k^{\mathrm{th}}$ row of $\bH$ together through a common scale parameter $\lambda_k$. If $\lambda_k$ is driven to zero (or, as we will see, a positive lower bound) during inference, then all entries in the  corresponding column  of $\bW$ and row of $\bH$ will also be driven to zero. 

\subsection{Priors} \label{sec:priors}

We are inspired by Bayesian PCA~\cite{bis99} where each element of $\bW$ is assigned a Gaussian prior with column-dependent variance-like parameters  $\lambda_k$. These $\lambda_k$'s are known as the  \emph{relevance weights}. However,  our formulation has two main differences vis-\`{a}-vis Bayesian PCA. Firstly, there are no nonnegativity constraints in Bayesian PCA. Secondly, in Bayesian PCA, thanks to the simplicity of the statistical model (multivariate Gaussian observations with Gaussian parameter priors), $\bH$ can be easily integrated out of the likelihood, and  the optimization can be done over $p(\bW, \blambda|\bV)$, where $\blambda = (\lambda_1,\ldots, \lambda_K)\in \bbR_+^K$ is the vector of relevance weights.  We have to maintain the nonnegativity of the elements in $\bW$ and $\bH$ and also in our setting the activation matrix $\bH$ cannot be integrated out analytically. 

To ameliorate the above-mentioned problems, we propose to tie the columns of $\bW$ and the rows of $\bH$ together through   common scale parameters. This construction is not over-constraining the scales of $\bW$ and $\bH$, because of the inherent scale indeterminacy between $\bw_k$ and $\underline{h}_k$.   Moreover, we choose nonnegative priors for $\bW$ and $\bH$ to ensure that all elements of the basis and activation matrices are nonnegative.  We adopt a Bayesian approach and assign  $\bW$ and $\bH$ Half-Normal or Exponential priors. When $\bW$ and $\bH$ have Half-Normal priors, 
\begin{align}
p(w_{fk}|\lambda_k)  = \calH\calN( w_{fk}| \lambda_k),\quad p(h_{kn}|\lambda_k)  = \calH\calN( w_{kn}| \lambda_k),\label{eqn:hn_prior}
\end{align}
where  for $x\ge 0$, $\calH\calN( x| \lambda ) \defeq (\frac{2}{\pi \lambda })^{1/2}  \exp(-\frac{x^2}{2\lambda } ),$ and $\calH\calN( x| \lambda )=0$ when $x<0$. Note that if $x$ is a Gaussian (Normal) random variable,  then $|x|$ is a Half-Normal. When $\bW$ and $\bH$ are assigned Exponential priors, 
\begin{align}
p(w_{fk}|\lambda_k) = \calE( w_{fk}|  \lambda_k) ,\quad p(h_{kn}|\lambda_k) = \calE( w_{kn}|  \lambda_k),\label{eqn:exp_prior}
\end{align}
where for $x\ge 0$, $\calE( x| \lambda )\defeq \frac{1}{\lambda}\exp (- \frac{x}{\lambda} )$, and $\calE( x| \lambda )=0$ otherwise. Note from~\eqref{eqn:hn_prior} and~\eqref{eqn:exp_prior} that the $k^{\mathrm{th}}$ column of $\bW$ and the $k^{\mathrm{th}}$ row of $\bH$ are tied together by a {\em common} variance-like  parameter $\lambda_k$, also known as the {\em relevance weight}. When a particular $\lambda_k$ is small, that particular column of $\bW$ {\em and} row of $\bH$  are not relevant and vice versa. When a row and a column are not relevant, their norms  are close to zero and thus can be removed from the factorization without compromising too much on data fidelity. This removal of {\em common} rows and columns makes the model more parsimonious.

Finally, we  impose   inverse-Gamma priors on each relevance weight $\lambda_k$, i.e., 
\begin{eqnarray}
p(\lambda_k; a,b) = \calI\calG(\lambda_k|a,b) = \frac{b^a}{\Gamma(a)}\lambda_k^{-(a+1)}\exp\left(-\frac{b}{\lambda_k}\right) \label{eqn:gamma_prior}
\end{eqnarray}
where $a$ and $b$ are  the (nonnegative) shape     and scale hyperparameters respectively.  We set $a$ and $b$ to be constant for all $k$. We will state how to choose these in a principled manner in Section~\ref{sec:selectab}. Furthermore, each relevance parameter is independent of every other, i.e., $p(\blambda; a,b) = \prod_{k=1}^K p(\lambda_k;a,b)$. 

\subsection{Likelihood} \label{sec:likel}

The $\beta$-divergence is related to the family of Tweedie distributions \cite{twee84}. The relation was noted by Cichocki {\it et al.} \cite{cicnmfbook} and detailed in \cite{yilm12}. The Tweedie distribution is a special case of the exponential dispersion model \cite{jor87}, itself a generalization of the more familiar natural exponential family. It is characterized by the  simple polynomial relation between its mean and variance
\begin{equation}
\var [x] = \phi \: \mu^{2-\beta}  \label{eqn:poly},
\end{equation}
where $\mu = \bbE[x]$ is the  mean,   $\beta$ is the {\em shape parameter} , and $\phi$ is referred to as the \emph{dispersion parameter}. The Tweedie distribution is only defined for $\beta \le 1$ and $\beta \ge 2$. For $\beta \not= 0, 1$, its probability density function (pdf) or probability mass function (pmf) can be written in the following form
\begin{eqnarray}
\mathcal{T}(x|\mu, \phi, \beta) = h(x,\phi) \exp \left[ \frac{1}{\phi} \left( \frac{1}{\beta-1} x \mu^{\beta-1} - \frac{1}{\beta} \mu^\beta \right) \right] \label{eqn:calT}
\end{eqnarray}
where $h(x,\phi)$ is referred to as the \emph{base function}. For $\beta \in \{0,1\}$,  the pdf or pmf takes the appropriate limiting form of~\eqref{eqn:calT}.  
The  support of $\mathcal{T}(x|\mu, \phi, \beta)$ varies with the value of $\beta$, but the set of values that $\mu$ can take on  is generally $\mathbb{R}^+$, except for $\beta = 2$, for which it is $\mathbb{R}$ and the Tweedie distribution coincides with the Gaussian distribution of mean $\mu$ and variance $\phi$. For $\beta = 1$ (and $\phi = 1$), the Tweedie distribution coincides with the Poisson distribution. For $\beta = 0$, it coincides with the Gamma distribution with shape parameter $\alpha = 1/\phi$ and scale parameter $\mu/\alpha$.\footnote{We employ the following convention for the Gamma distribution $\calG(x;a,b) =x^{a-1} e^{-x/b}/(b^a\Gamma(a))$.} The base function admits a closed form only  for  $\beta \in \{-1,0,1,2\}$. 

Finally, the \emph{deviance} of Tweedie distribution, i.e., the log-likelihood ratio of  the saturated ($\mu = x$) and general model is proportional to the $\beta$-divergence. In particular,
\bal{ \label{eqn:likelihood1}
\log \frac{ \mathcal{T}(x  |\mu = x, \phi, \beta) }{  \mathcal{T}(x  |\mu, \phi, \beta)} = \frac{1}{\phi} d_{\beta}(x|\mu),
}
where $d_{\beta}(\fndot |\fndot)$ is  the scalar cost function defined in  \eqref{eqn:beta}. As such the $\beta$-divergence acts as a minus log-likelihood for the Tweedie distribution, whenever the latter is defined.
Because the data coefficients $\{ v_{fn} \}$  are conditionally independent given $(\bW, \bH)$,  the negative log-likelihood function is
\begin{equation}
-\log p(\bV|\bW,\bH)= \frac{1}{\phi} D_\beta(\bV|\bW\bH)+\cst . \label{eqn:likelihood}
\end{equation}

\subsection{Objective function}

We now form the maximum a posteriori (MAP) objective function for the model   described in Sections~\ref{sec:priors} and~\ref{sec:likel}. On account of~\eqref{eqn:hn_prior}, \eqref{eqn:exp_prior},  \eqref{eqn:gamma_prior}  and \eqref{eqn:likelihood},
\begin{align}
& C(\bW,   \bH,\blambda)  \defeq -\log p(\bW,\bH,\blambda|\bV)\label{eqn:cost1}  \\
&= \frac{1}{\phi}  D_\beta(\bV|\bW\bH)   +   \sum_{k=1}^K    \frac{1}{\lambda_k} \left( f(\bw_k) + f(\underline{h}_k) + b \right)    \nonumber\\*
&\hspace{1.5in}   +    c \log \lambda_k   + \cst ,  \label{eqn:cost3}
\end{align}
where \eqref{eqn:cost3} follows from Bayes' rule and for the two  statistical models, 
\begin{itemize}
\item Half-Normal model as in~\eqref{eqn:hn_prior},  $f(\bx) = \frac{1}{2}\|\bx\|_2^2$ and $c={(F+N)}/{2} + a+1$. 
\item Exponential model  as in~\eqref{eqn:exp_prior},  $f(\bx) = \|\bx\|_1$ and $c= F+N+a+1$. 
\end{itemize}

Observe that for the regularized cost function in \eqref{eqn:cost3},    the second term is monotonically decreasing in $\lambda_k$ while the third term is monotonically increasing in $\lambda_k$. Thus, a subset of the $\lambda_k$'s will be forced to a lower bound which we specify in Section~\ref{sec:stopping} while the others will tend to a larger value. This serves the purpose of pruning   irrelevant components out of the model. In fact, the vector of relevance parameters $\blambda = (\lambda_1, \ldots, \lambda_K)$ can be optimized analytically in~\eqref{eqn:cost3} leading to an objective function that is a function of $\bW$ and $\bH$ only, i.e.,  
\begin{multline} \label{eqn:lasso_cost}
C(\bW,\bH)= \\
 \frac{1}{\phi}  D_\beta(\bV|\bW\bH) + c   \sum_{k=1}^K   \log \left( f(\bw_k)  \! +\!  f(\underline{h}_k)  \! + \!b \right) + \cst , 
\end{multline}
where $\cst = K c (1-\log c)$. 

In our algorithms,  instead of optimizing \eqref{eqn:lasso_cost}, we  keep $\lambda_k$ as an auxiliary variable for   optimizing $C(\bW,\bH,\blambda)$ in~\eqref{eqn:cost3} to ensure that the columns $\bH$ and the rows of $\bW$ are decoupled. More precisely, $\bw_k$ and $\underline{h}_k$ are conditionally independent given $\lambda_k$.  In fact,  \eqref{eqn:lasso_cost} shows that the $\blambda$-optimized objective function $C(\bW,\bH)$ induces sparse regularization among  groups, where the groups are pairs of columns and rows, i.e., $\{\ve{w}_k, \underline{h}_k\}$. In this sense, our work is related to group LASSO~\cite{yuan} and its variants. See for example~\cite{bach12}. The function $x\mapsto\log(x+b)$   in \eqref{eqn:lasso_cost} is a sparsity-inducing term  and is related to reweighted $\ell_1$-minimization~\cite{CandesReweighted}. We discuss these connections in greater detail  in the supplementary material~\cite{suppMat}.

\section{Inference Algorithms} \label{sec:algos}
In this section, we describe two  algorithms  for optimizing the objective function~\eqref{eqn:cost3} for $\bH$ given fixed  $\bW$.  The updates for $\bW$ are symmetric given $\bH$. These algorithms will be based on the MM idea for $\beta$-NMF and  on the two prior distributions of $\bW$ and $\bH$. In particular, we use the auxiliary function $G(\bH|\tilde{\bH})$ defined in Table~\ref{tab:aux} as an upper bound of the data fit term $D_\beta(\bV|\bW\bH)$. 

\subsection{Algorithm for $\ell_2$-ARD $\beta$-NMF} \label{sec:l2}
We now introduce  $\ell_2$-ARD $\beta$-NMF. In this algorithm, we assume that $\bW$ and $\bH$ have Half-Normal priors as in \eqref{eqn:hn_prior} and thus, the regularizer   is 
\begin{equation} \label{eqn:reg_l2}
R_2(\bH)\defeq \sum_{k} \frac{1}{\lambda_k} f(\underline{h}_k)=\sum_{kn}\frac{1}{2\lambda_k}  h_{kn}^2.
\end{equation}
The main idea behind the  algorithms is as follows: Consider the function  $F(\bH|\tilde{\bH})\defeq \phi^{{-1}} \, G(\bH|\tilde{\bH})+ R_2(\bH)$ which is the original auxiliary function $G(\bH|\tilde{\bH})$ times $\phi^{-1}$ plus the $\ell_2$ regularization term. It can in fact be easily shown in~\cite[Sec.~6]{betanmf} that $F(\bH|\tilde{\bH})$ is an auxiliary function to the  (penalized) objective  function in~\eqref{eqn:cost3}.  Ideally, we would take the derivative of $F(\bH|\tilde{\bH})$ w.r.t\ $h_{kn}$ and  set it to zero. Then the updates would proceed in a manner analogous to~\eqref{eqn:update_hkn}. However, the regularization term   $R_2(\bH)$ does not ``fit well'' with the form of the auxiliary function $G(\bH|\tilde{\bH})$ in the sense that $\nabla_{\bH}F(\bH|\tilde{\bH})=0$ cannot be solved analytically for all $\beta \in\bbR$. Thus, our idea for $\ell_2$-ARD is to consider the cases $\beta\ge 2$ and $\beta<2$ separately and  to find an upper bound  of  $F(\bH|\tilde{\bH})$ by some other auxiliary function $\tilF(\bH|\tilde{\bH})$ so that the resulting equation $\nabla_{\bH}\tilF(\bH|\tilde{\bH})=0$ can be solved in closed-form. 

To derive our algorithms, we first note the following:
\begin{lemma}\label{lem:incr}
For every ${\nu}>0$, the function $g_\nu(t)=\frac{1}{t}({\nu}^t-1)$ is monotonically non-decreasing in $t\in\bbR$. In fact, $g_\nu(t)$ is monotonically  increasing unless $\nu=1$.
\end{lemma} 
In the above lemma,   $g_\nu(0) \defeq  \log {\nu}$ by L'H\^{o}pital's rule. The proof of this simple result can be found in \cite{yang11}.

We first derive $\ell_2$-ARD for $\beta> 2$. The idea is to upper bound the regularizer $R_2(\bH)$  in~\eqref{eqn:reg_l2} elementwise using  Lemma~\ref{lem:incr}, and is equivalent to the \emph{moving-term} technique described by Yang and Oja in~\cite{yang10b, yang11}. Indeed, we have 
\begin{equation}
\frac{1}{2} \left[ \left(\frac{h_{kn}}{\tilh_{kn}} \right)^2-1\right]\le \frac{1}{\beta} \left[ \left(\frac{h_{kn}}{\tilh_{kn}} \right)^\beta-1\right] 
\end{equation}
by taking ${\nu}=h_{kn}/\tilh_{kn}$ in Lemma~\ref{lem:incr}. Thus, for $\beta>2$,
\begin{equation}\label{eqn:upper_bound_l1}
\frac{1}{2\lambda_k} h_{kn}^2\le \frac{1}{\lambda_k\beta} \tilh_{kn}^2 \left(\frac{h_{kn}}{\tilh_{kn}} \right)^{\beta} +\cst
\end{equation}
where $\cst$ is a constant w.r.t the optimization variable $h_{kn}$. We upper bound the  regularizer~\eqref{eqn:reg_l2} elementwise   by~\eqref{eqn:upper_bound_l1}. The resulting auxiliary function (modified version of $F(\bH|\tilde{\bH})$) is  
\begin{equation}
\tilF(\bH|\tilde{\bH})= \frac{1}{\phi} \, G(\bH|\tilde{\bH})+\sum_{kn} \frac{1}{\lambda_k\beta} \tilh_{kn}^2 \left(\frac{h_{kn}}{\tilh_{kn}} \right)^{\beta}.
\end{equation}
Note that \eqref{eqn:upper_bound_l1} holds with equality iff ${\nu}=1$ or equivalently,  $h_{kn}=\tilh_{kn}$ so~\eqref{eqn:eqaulity} holds. Thus, $\tilF(\bH|\tilde{\bH})$ is indeed  an auxiliary function to $F(\bH|\tilde{\bH})$.  Recalling the definition of $G(\bH|\tilde{\bH})$ for $\beta>2$ in Table~\ref{tab:aux}, differentiating $\tilF(\bH|\tilde{\bH})$ w.r.t\ $h_{kn}$ and setting the result to zero yields
the update
\begin{equation}  
h_{kn} = \tilde{h}_{kn} \left( \frac{p_{kn}}{q_{kn}+ (\phi/\lambda_k) \tilh_{kn}} \right)^{1/(\beta-1)}. \label{eqn:updatel21}
\end{equation}
Note that the exponent $1/(\beta-1)$ corresponds to $\gamma(\beta)$ for the $\beta>2$ case. Also observe that the update is similar to MM for $\beta$-NMF [cf.~\eqref{eqn:update_hkn}] except that there is an additional term in the denominator.

\begin{algorithm}[t]
\caption{$\ell_2$-ARD for $\beta$-NMF}
\begin{algorithmic}\label{alg:l2}
\STATE {\bf Input:}   Data matrix $\bV$,    hyperparameter $a$, tolerance $\tau$
\STATE {\bf Output:}  Nonnegative matrices $\bW$ and $\bH$,  nonnegative relevance  vector $\blambda$ and  model order  $\Keff$ 
\STATE {\bf Init:}  Fix $K$. Initialize $\bW \in \bbR_+^{F\times K}$ and $\bH\in \bbR_+^{K\times N}$ to nonnegative values and tolerance parameter tol $=\infty$
\STATE {\bf Calculate:} $c=(F+N)/2+a+1$ and   $\xi(\beta)$ as in \eqref{eqn:xi}
\STATE {\bf Calculate:} Hyperparameter $b$ as in \eqref{eqn:choose_b}
\WHILE {(tol $<\tau$)}
\STATE   $\bH \leftarrow \bH \cdot \left( \frac{{\bW^T [(\bW\bH)^{\cdot(\beta-2)} \cdot \bV]} }{{\bW^T [(\bW\bH)^{\cdot(\beta-1)}  ]} + \phi \, \bH/\repmat(\blambda,1,N) }\right)^{\cdot \, \xi(\beta)}$
\STATE   $\bW \leftarrow \bW \cdot \left( \frac{{[(\bW\bH)^{\cdot(\beta-2)} \cdot \bV]\bH^T} }{{[(\bW\bH)^{\cdot(\beta-1)}  ] \bH^T} + \phi \, \bW/\repmat(\blambda,F,1) }\right)^{\cdot \, \xi(\beta)}$
\STATE   $\lambda_k \leftarrow [(\frac{1}{2} \sum_f w_{fk}^2 +  \frac{1}{2} \sum_n h_{kn}^2)  +  b]/c$ for all $k$
\STATE   tol $\leftarrow$ $\max_{k=1,\ldots, K} | (\lambda_k-\tilde{\lambda}_k)/ \tilde{\lambda}_k |$
\ENDWHILE
\STATE {\bf Calculate:} {$\Keff$ as in \eqref{eqn:Keff}}
\end{algorithmic}
\end{algorithm}

For the case $\beta\le 2$, our strategy is not to majorize the regularization term. Rather we majorize the the auxiliary function $G(\bH|\tilde{\bH})$ itself. By applying Lemma~\ref{lem:incr} with ${\nu}=h_{kn}/\tilh_{kn}$, we have that for all $\beta\le 2$, 
\begin{equation}
 \frac{1}{\beta} \left[ \left(\frac{h_{kn}}{\tilh_{kn}} \right)^\beta-1\right]\le\frac{1}{2} \left[ \left(\frac{h_{kn}}{\tilh_{kn}} \right)^2-1\right]
\end{equation}
which means that 
\begin{equation}
\frac{1}{\beta} q_{kn} \tilh_{kn} \left(\frac{h_{kn}}{\tilh_{kn}} \right)^\beta \le \frac{1}{2}q_{kn} \tilh_{kn}  \left(\frac{h_{kn}}{\tilh_{kn}} \right)^2+\cst .
\end{equation}
By replacing the first term of $G(\bH|\tilde{\bH})$  in Table~\ref{tab:aux}  (for $\beta\le 2$) with the upper bound above, we have the following new objective function
\begin{align}
\tilF(\bH|\tilde{\bH})&=  \sum_{kn} \frac{ \, q_{kn} \tilde{h}_{kn}}{2 \phi}    \left( \frac{h_{kn}}{\tilde{h}_{kn}} \right)^{2}     \nn\\
&\qquad\quad -\frac{  p_{kn} \tilde{h}_{kn}}{\phi (\beta-1)} \left( \frac{h_{kn}}{\tilde{h}_{kn}} \right)^{\beta-1}  \!  + \frac{h_{kn}^2}{2\lambda_k} .
\end{align}
Differentiating $\tilF(\bH|\tilde{\bH})$ w.r.t\ $h_{kn}$ and setting to zero yields
the simple update
\begin{equation}
h_{kn} = \tilde{h}_{kn} \left( \frac{p_{kn}}{q_{kn}+ (\phi/\lambda_k)\tilh_{kn}} \right)^{1/(3-\beta)}. \label{eqn:updatel22}
\end{equation}
To summarize the algorithm concisely, we define the exponent used in the updates in~\eqref{eqn:updatel21} and~\eqref{eqn:updatel22} as
\begin{equation} \label{eqn:xi}
\xi(\beta) \defeq \left\{ \begin{array}{cr}
1/(3-\beta) & \beta\le 2 \\
1/(\beta-1) & \beta > 2
\end{array}  \right. .
\end{equation}
Finally, we remark that even though the updates in~\eqref{eqn:updatel21} and~\eqref{eqn:updatel22} are easy to implement,   we either majorized the regularizer $R_2(\bH)$ or the   auxiliary function $G(\bH|\tilde{\bH})$. These bounds may be loose and thus may lead to slow convergence in  the resulting algorithm. In fact, we can show that for $\beta=0,1,2$, we do not have to resort to upper bounding the original function $F(\bH|\tilde{\bH})={\phi^{-1}} \, G(\bH|\tilde{\bH}) +R_2(\bH)$. Instead, we can choose to solve a polynomial equation to update $h_{kn}$. The cases $\beta=0,1,2$ correspond respectively to solving cubic, quadratic and linear equations in $h_{kn}$ respectively.  It is also true that for all rational $\beta$, we can form a polynomial equation in $h_{kn}$ but the order of the resulting polynomial depends on the exact value of $\beta$. See the supplementary material~\cite{suppMat}. 

\subsection{Algorithm for $\ell_1$-ARD $\beta$-NMF} \label{sec:l1}
The derivation of  $\ell_1$-ARD $\beta$-NMF is   similar to its $\ell_2$ counterpart. We find majorizers for either the likelihood or the regularizer.  We omit the derivations and refer the reader to the supplementary material~\cite{suppMat}. In sum, 
\begin{equation}
h_{kn} = \tilde{h}_{kn} \left( \frac{p_{kn}}{q_{kn}+ \phi/\lambda_k} \right)^{\gamma(\beta)}, \label{eqn:updatel12}
\end{equation}
where $\gamma(\beta)$ is defined in \eqref{eqn:defgamma}.

\subsection{Update of $\lambda_k$} \label{sec:lambda}
We have described how to update $\bH$ using either $\ell_1$-ARD or $\ell_2$-ARD. Since $\bH$ and $\bW$ are related in a  symmetric manner, we have also effectively described how to update $\bW$.  We now describe a simple update rule for the $\lambda_k$'s.  This update is the same for both  $\ell_1$- and $\ell_2$-ARD. We   first find  the partial derivative of $C(\bW,\bH,\blambda)$  w.r.t\ $\lambda_k$ and set it to zero. This gives the update:
\begin{equation}
\lambda_k =  \frac{f(\bw_k)+f(\underline{h}_k)+b}{c}, \label{eqn:lambda_update}
\end{equation}
where $f(\fndot)$ and $c$ are defined after \eqref{eqn:cost3}.

\begin{algorithm}[t]
\caption{$\ell_1$-ARD for $\beta$-NMF}
\begin{algorithmic} \label{alg:l1}
\STATE {\bf Input:}   Data matrix $\bV$,    hyperparameter $a$, tolerance $\tau$
\STATE {\bf Output:}  Nonnegative matrices $\bW$ and $\bH$,  nonnegative relevance  vector $\blambda$ and  model order  $\Keff$ 
\STATE {\bf Init:}  Fix $K$. Initialize $\bW \in \bbR_+^{F\times K}$ and $\bH\in \bbR_+^{K\times N}$ to nonnegative values and tolerance parameter tol $=\infty$
\STATE {\bf Calculate:} $c=F+N+a+1$ and   $\gamma(\beta)$ as in \eqref{eqn:defgamma}
\STATE {\bf Calculate:} Hyperparameter $b$ as in \eqref{eqn:choose_b}
\WHILE {(tol $<\tau$)}
\STATE  $\bH  \leftarrow \bH\cdot \left( \frac{{ \bW^T [(\bW\bH)^{\cdot(\beta-2)} \cdot \bV] } }{{\bW^T [(\bW\bH)^{\cdot(\beta-1)}  ]} + \phi/\repmat(\blambda,1,N) }\right)^{\cdot\, \gamma(\beta)}$
\STATE   $\bW   \leftarrow \bW \cdot\left( \frac{{[(\bW\bH)^{\cdot(\beta-2)} \cdot \bV]\bH^T} }{{[(\bW\bH)^{\cdot(\beta-1)}  ] \bH^T} + \phi/\repmat( \blambda,F,1) }\right)^{\cdot\, \gamma(\beta)}$
\STATE   $\lambda_k \leftarrow ( \sum_f w_{fk}+\sum_n h_{kn} +b)/c$ for all $k$
\STATE    tol $\leftarrow$ $\max_{k=1,\ldots, K} | (\lambda_k-\tilde{\lambda}_k)/  \tilde{\lambda}_k  |$ 
\ENDWHILE
\STATE {\bf Calculate:}  {$\Keff$ as in \eqref{eqn:Keff}}
\end{algorithmic}
\end{algorithm}

\subsection{Stopping criterion and determination of $\Keff$} \label{sec:stopping}
In this section, we describe the stopping criterion and the determination of the effective number of components $\Keff$. Let $\blambda = (\lambda_1,\ldots, \lambda_K)$ and $\tilde{\blambda} = (\tilde{\lambda}_1,\ldots, \tilde{\lambda}_K)$ be the vector of relevance weights at the current (updated) and previous iterations respectively. The algorithm is terminated whenever
$
\mathrm{tol} \defeq \max_{k=1,\ldots, K} | {(\lambda_k-\tilde{\lambda}_k)} / {\tilde{\lambda}_k} | 
$
falls below some threshold $\tau>0$. Note from \eqref{eqn:lambda_update} that iterates of each $\lambda_k$   are bounded from below as
$
\lambda_k \ge B\defeq b/c$ and this bound is attained when $\bw_k$ and $\underline{h}_k$ are zero vectors, i.e., the $k^{\mathrm{th}}$ column of $\bW$ and the $k^{\mathrm{th}}$  row of $\bH$ are pruned out of the model. After convergence, we set $\Keff$  to be the number of components of such that the ratio $(\lambda_k-B)/B$ is strictly larger than $\tau$, i.e., 
\begin{equation}
\Keff \defeq \left|\left\{   k\in\{1,\ldots , K\} :  \frac{\lambda_k-B}{B}>\tau \right\} \right|, \label{eqn:Keff}
\end{equation}
where $\tau>0$ is  some threshold. We choose this threshold to be the same  as that for the tolerance level $\mathrm{tol}$.

The algorithms $\ell_2$-ARD and $\ell_1$-ARD  are detailed in Algorithms~\ref{alg:l2} and~\ref{alg:l1} respectively. In the algorithms, we use the notation $\bA\cdot \bB$ to mean entrywise multiplication   of $\bA$ and $\bB$; $\frac{\bA}{\bB}$ to mean entrywise division; and $\bA^{\cdot\, \gamma}$ to mean entrywise raising to the $\gamma^{\mathrm{th}}$ power. In addition, $\repmat(\blambda, 1,N)$ denotes the $K\times N$ matrix with each column being   the $\blambda$ vector.

\subsection{Choosing the hyperparameters} \label{sec:selectab}

\subsubsection{Choice of dispersion parameter $\phi$} \label{sec:disp}
The dispersion parameter $\phi$ represents the tradeoff between the data fidelity   and the regularization terms in~\eqref{eqn:cost3}. It needs to be    fixed, based on prior knowledge about the noise distribution, or    learned from the data using either cross-validation or MAP estimation. In the latter case, $\phi$ is assigned a prior $p(\phi)$ and the objective  $C(\bW,\bH,\lambda,\phi)$ can be optimized over $\phi$.   This is a standard feature in  penalized likelihood approaches and has been widely discussed in the literature.  In  this work, we will not  address the estimation of $\phi$, but only study the influence the regularization term on the factorization \emph{given $\phi$}.  In many cases, it is reasonable to fix $\phi$ based on prior knowledge. In particular, under the Gaussian noise assumption, 
$
v_{fn} \sim \mathcal{N}(v_{{fn}}|\hat{v}_{fn},\sigma^2),
$
 and $\beta=2$ and $\phi=\sigma^2$. Under the Poisson noise assumption, 
$
v_{fn} \sim \mathcal{P}(v_{{fn}}|\hat{v}_{fn}),
$
 and  $\beta =1$ and $\phi = 1$. Under multiplicative Gamma noise assumption,  $v_{fn} = \hat{v}_{fn}\cdot \epsilon_{fn}$ and $\epsilon_{fn}$ is a Gamma noise of mean 1, or equivalently
$
v_{fn} \sim \calG(v_{kn} | \alpha , \hatv_{fn}/ \alpha ),
$  
 and $\beta = 0$ and $\phi = 1/\alpha$. In audio applications where the power spectrogram is to be factorized, as in Section~\ref{sec:audio}, the multiplicative exponential noise model (with $\alpha  = 1$) is a generally agreed upon assumption \cite{neco09} and thus $\phi = 1$.  


\subsubsection{Choice of hyperparameters $a$ and $b$} \label{sec:chooseab}

We now discuss how to  choose  the hyperparameters $a$ and $b$ in~\eqref{eqn:gamma_prior} in a data-dependent and principled way. Our method is related to the  {\em method of moments}. We first focus on the selection of $b$ using the sample mean of data, given $a$. Then the selection of $a$ based on the sample variance of the data is  discussed at the end of the section.

Consider the approximation in \eqref{eqn:facto}, which can be written element-wise as 
\begin{equation}
v_{fn}\approx \hatv_{fn}=\sum_k w_{fk} h_{kn}. \label{eqn:element-wise}
\end{equation}
The statistical models corresponding to shape parameter $\beta \notin (1,2)$ imply that $\bbE [v_{fn}|\hatv_{fn} ]=\hatv_{fn}$.   We extrapolate this   property to derive a rule for selecting the hyperparameter $b$ for all $\beta\in\bbR$ (and for nonnegative real-valued data in general) even though there is no  known statistical model governing the noise when $\beta \in (1,2)$. When $F N$ is large, the law of large numbers implies that the sample mean of the elements in  $\bV$ is close to the population mean (with high probability), i.e., 
\begin{eqnarray}
\hmu_{\bV} \defeq\frac{1}{FN}\sum_{fn}v_{fn} \approx\bbE [v_{fn}  ]=\bbE [\hatv_{fn} ] = \sum_k \bbE [w_{fk} h_{kn} ]
\end{eqnarray}
We can compute $\bbE [\hatv_{fn} ]$ for the Half-Normal and Exponential models using the moments of these distributions and those of the inverse-Gamma for $\lambda_k$. These   yield
\begin{align}
\bbE[\hatv_{fn} ] = \left\{   \begin{array}{cc}
\frac{2K b}{\pi(a-1)} & \mbox{Half-Normal} \\
\frac{K b^2}{(a-1)(a-2)} & \mbox{Exponential}
\end{array}
\right. . \label{eqn:expectation}
\end{align}
By equating these expressions to the empirical mean $\hmu_{\bV}$, we conclude that we can choose $b$ according to 
\begin{align}
\hatb  = \left\{   \begin{array}{cc}
\frac{\pi (a-1)  \hmu_{\bV}}{2 K } &  \ell_2\mbox{-ARD} \\
 \sqrt{  \frac{(a-1)(a-2)  \hmu_{\bV} }{K}   }&  \ell_1\mbox{-ARD}
\end{array}
\right. . \label{eqn:choose_b}
\end{align}
In summary, $\hatb \propto  \hmu_{\bV}/K$ and $\hatb \propto (\hmu_{\bV}/K)^{1/2}$ for   $\ell_2$- and $\ell_1$-ARD  respectively. 

By using the empirical variance of $\bV$ and the relation between the mean and variance of the Tweedie distribution in~\eqref{eqn:poly}, we may also estimate $a$ from the data. The resulting relations are more involved and these calculations are deferred to the supplementary~material~\cite{suppMat}  for $\beta \in \{ 0, 1, 2 \}$. However, experiments showed that the resulting learning rules for $a$ did not consistently give satisfactory results, especially when $FN$ is not sufficiently large. In particular, the estimates sometimes fall out of the parameter space, which is a known feature of the method of moments. Observe that $a$ appears in the objective function~\eqref{eqn:lasso_cost} only through $c = (F+N)/2 + a +1$ ($\ell_2$-ARD) or $c = F+N + a +1$ ($\ell_1$-ARD). As such, the influence of $a$ is moderated by $F+N$. Hence, if we want to choose a prior on $a$ that is not too informative, then we should choose $a$ to be small compared to $F+N$. Experiments in Section~\ref{sec:exps} confirm that smaller values of $a$ (relative to $F+N$) typically produce better  results. As discussed in the conclusion, a more robust estimation of $a$ (as well as $b$ and $\phi$) would involve a fully Bayesian treatment of our problem, which is left for future work.

\section{Connections with other works}  \label{sec:connect}
Our work draws parallel with a few other works on model order selection in NMF. The closest work is~\cite{mor09} which also proposes automatic component pruning via a MAP approach. It was developed during the same period as and independently of our earlier work~\cite{spars09}. An extension to multi-array analysis is also proposed in~\cite{mor09b}. In~\cite{mor09}, M{\o}rup \& Hansen consider NMF with the Euclidean and KL costs. They constrained the columns of $\bW$ to have unit norm (i.e., $\| \ve{w}_k \|_2 = 1$) and assumed that the coefficients of $\bH$ are assigned  exponential priors $\calE(h_{kn} | \lambda_k)$. A non-informative Jeffrey's prior is further  assumed on $\lambda_k$. Put together, they consider the following  optimization over   $(\bW,\bH)$:
\begin{align}
& \mathop{\mathrm{minimize}}_{\bW,\bH,\blambda} \quad  \ D(\bV | \bW \bH) + \sum_k \frac{1}{\lambda_k} \| \underline{h}_k \|_1 + N \log \lambda_k \nn\\ 
& \st \quad \bW \ge 0, \,\,  \bH \ge 0, \,\, \|\ve{w}_k\|_2 = 1, \,\, \forall\, k\label{eqn:morup2}
\end{align}
where $D(\cdot|\cdot)$ is either the squared Euclidean distance or the KL-divergence. A major difference compared to our objective function in~\eqref{eqn:cost3} is that this method involves optimizing $\bW$ under the constraint $\|\ve{w}_k\|_2 = 1$, which is non-trivial. As such, to solve~\eqref{eqn:morup2} the authors in~\cite{mor09} use a change of variable $\ve{w}_k'\leftarrow \ve{w}_k/\|\ve{w}_k\|_2$ and derive a heuristic multiplicative algorithm based on the ratio of negative and positive parts of the new  objective function, along the lines of~\cite{egg04}. In contrast, our approach treats $\ve{w}_k$ and $\underline{h}_k$ symmetrically and the updates are simple. Furthermore, the pruning approach  in~\cite{mor09} only occurs in the rows $\bH$ and the corresponding columns of $\bW$ may take any nonnegative value (subject to the norm constraint), which makes the estimation of these columns of $\bW$ ill-posed (i.e., the parametrization is such that a part of the model is not observable). In contrast, in our approach $\ve{w}_k$ and $\underline{h}_k$ are tied together so they converge to zero jointly when $\lambda_k$ reaches its lower bound. 

Our work is also related to the automatic rank determination method in Projective NMF proposed by Yang et al.\ in~\cite{yang10}. Following the principle of  PCA, Projective NMF seeks a nonnegative matrix $\bW$ such that the projection of $\bV$ on the subspace spanned by $\bW$ best fits $\bV$. In other words, it is assumed that $\bH = \bW^T \bV$. Following ARD in Bayesian PCA as originally described by Bishop~\cite{bis99}, Yang et al.\ consider the additive Gaussian noise model and propose to place half-normal priors with relevance parameters $\lambda_k$ on the columns of $\bW$. They describe how to adapt  EM to achieve MAP estimation of $\bW$ and its relevance parameters.

Estimation of the model order in the Itakura-Saito NMF (multiplicative exponential noise) was addressed by Hoffman et al.~\cite{hoffman}. They employ a nonparametric Bayesian setting in which $K$ is assigned a large value (in principle, infinite) but the model is such that only a finite subset of components is retained. In their model, the coefficients of $\bW$ and $\bH$ have Gamma priors with fixed hyperparameters and a weight parameter $\theta_k$ is placed before each component in the factor model, i.e., $\hat{v}_{fn} = \sum_k \theta_k w_{fk} h_{kn}$. The weight, akin to the relevance parameter in our setting, is assigned a Gamma prior with a sparsity-enforcing shape parameter. A difference with our model  is the {\em a priori} independence of the factors and the weights. Variational inference is used in \cite{hoffman}. 

In contrast with the above-mentioned works, the work herein presents a unified framework for model selection in $\beta$-NMF. The proposed algorithms have low complexity per iteration and are simple to implement, while decreasing the objective function at every iteration. We compare the performance of our algorithms to those in~\cite{mor09} and~\cite{hoffman} in Sections~\ref{sec:audio} (music decomposition) and \ref{eqn:stock} (stock price prediction).

\section{Experiments} \label{sec:exps}
In this section, we present extensive numerical experiments  demonstrating the robustness and efficiency of the proposed algorithms for (i) uncovering the correct model order and (ii) learning better decompositions for modeling nonnegative data. 
\subsection{Simulations with synthetic data} \label{sec:synth}
In this section, we describe   experiments on synthetic data generated according to the model.  In particular, we fixed a pair of hyperparameters $(a_{\mathrm{true}},b_{\mathrm{true}})$ and sampled   $K_{\mathrm{true}}=5$ relevance weights $\lambda_k$ according to the inverse-Gamma prior in~\eqref{eqn:gamma_prior}. Conditioned on these relevance weights, we sampled the elements of $\bW$ and $\bH$ from the Half-Normal or Exponential models depending on whether we chose to use $\ell_2$- or $\ell_1$-ARD. These models are defined in~\eqref{eqn:hn_prior} and~\eqref{eqn:exp_prior}  respectively.  We set $a_{\mathrm{true}}=50$ and $b_{\mathrm{true}}= 70$ for reasons that will be made clear in the following. We define the noiseless matrix $\hat{\bV}$ as  $\bW \bH$. We then generated a noisy matrix $\bV$ given $\hat{\bV}$ according to the three  statistical models  $\beta=0,1,2$ corresponding to IS-, KL-  and EUC-NMF respectively.  More precisely, the parameters of the noise models are chosen so that the signal-to-noise ratio $\SNR$ in dB, defined as 
$
\SNR = 20\log_{10}  ({\|\hat{\bV}\|_F  } / {\|\bV-\hat{\bV}\|_F }  ) , 
$
is approximately 10 dB   for  each $\beta\in\{0,1,2\}$.  For $\beta=0$, this corresponds to an $\alpha$, the shape parameter, of approximately $10$. For $\beta=1$, the parameterless Poisson noise model  results in an {\em integer-valued} noisy matrix $\bV$. Since there is no noise parameter to select Poisson noise model, we chose $b_{\mathrm{true}}$ so that the elements of the data matrix $\bV$ are  large enough resulting in an $\SNR\approx$ 10 dB. For the Gaussian observation model ($\beta=2$), we can analytically solve for the noise variance $\sigma^2$ so that the $\SNR$ is approximately 10 dB.  In addition, we set  the number of columns $N = 100$, the initial number of components $K=2 \, K_{\mathrm{true}}=10$ and chose two different values for $F$, namely 50 and 500. The threshold value $\tau$ is set to $10^{-7}$ (refer to Section~\ref{sec:stopping}). It was observed using this value of the threshold that the iterates of $\lambda_k$ converged to their limiting values.  We ran $\ell_1$- and $\ell_2$-ARD for  $a\in \{5,10,25,50,100, 250, 500\}$ and using $b$ computed as in Section~\ref{sec:chooseab}. The dispersion parameter $\phi$ is assumed known and set as in the discussion after~\eqref{eqn:likelihood}.   

\begin{figure}[t]
\begin{center}
   \includegraphics[width=\linewidth]{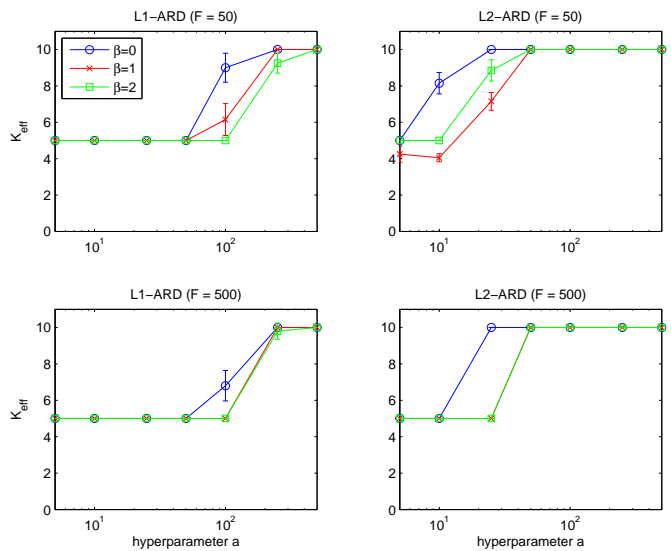}    
   \caption{Estimated number of components as a function  of  the hyperparameter $a$ (log-linear plot). The true model order is $K_{\mathrm{true}}=5$. The solid line is the mean across 10 runs and the error bars display $\pm$ the standard deviation.}
   \label{fig:syn} \end{center}
\end{figure}

To make fair comparisons, the data and the initializations are the same for $\ell_2$- and $\ell_1$-ARD as well as for every $(\beta,a)$.  We  averaged the inferred model order $\Keff$ over 10 different  runs. The results are displayed in Fig.~\ref{fig:syn}. 

Firstly,  we observe that $\ell_1$-ARD recovers the model order $K_{\mathrm{true}}=5$ correctly when $a\le 100$ and $\beta\in  \{0,1,2\}$. This range includes $a_{\mathrm{true}}=50$, which is the true hyperparameter we generated the data from. Thus, if we use the correct range of values of $a$, and if the $\SNR$ is of the order 10 dB (which is reasonable in most applications),  we are able to recover the true model order from the data. On the other hand, from the top right and bottom right plots, we see that $\ell_2$-ARD is not as robust in recovering the right latent dimensionality.

Secondly, note that the quality of estimation is relatively consistent across various $\beta$'s. The success of the proposed algorithms hinges more on the amount of noise added (i.e., the $\SNR$) compared to which specific $\beta$ is assumed. However, as discussed in Section~\ref{sec:likel}, the shape parameter $\beta$ should be chosen to reflect our belief in the underlying generative model and the noise statistics. 

Thirdly, observe that when more data are available ($F=500$), the estimation quality improves significantly. This is evidenced by the fact that even $\ell_2$-ARD (bottom right plot) performs much better -- it selects the right model order for all $a\le 25$ and $\beta \in  \{1,2\}$. The estimates are also   much more consistent across various initializations. Indeed the standard deviations for most sets of experiments is zero, demonstrating that there is little or no variability due to   random initializations. 
 
\subsection{Simulations with the  \texttt{swimmer} dataset} \label{sec:swim}

\begin{figure}[t]
 \begin{center}
   \includegraphics[width=\linewidth]{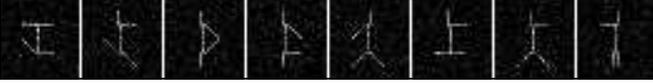}    
   \caption{Sample images of the noisy \texttt{swimmer} data. The colormap is adjusted such that black corresponds to the smallest data coefficient value ($v_{fn}=0$) and white the largest ($v_{fn}= 24$).}
   \label{fig:swim_data}
   \end{center}
\end{figure}

In this section we report experiments on the \texttt{swimmer} dataset introduced in~\cite{don04}. This is a synthetic dataset of $N=256$ images each of size $F= 32 \times 32 = 1024$. Each image represents a swimmer composed of an invariant torso and four limbs, where each limb can take one of four positions. We set background pixel values to 1 and body pixel values to 10, and generated noisy data with Poisson noise. Sample images of the resulting noisy data are shown in Fig.~\ref{fig:swim_data}. The ``ground truth'' number of components for this dataset is $K_{\mathrm{true}} = 16$, which corresponds to all the different limb positions. The torso and background form an invariant component that can be associated with any of the four limbs, or equally split among limbs. The data images are vectorized and arranged in the columns of $\ve{V}$.

\begin{figure}[t]
 \begin{center}
   \includegraphics[width=\linewidth]{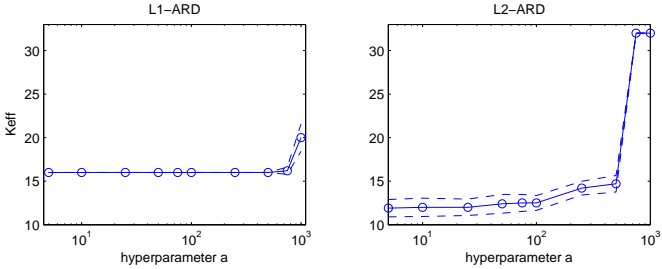}    
   \caption{Estimated number of components $K_\text{eff}$ as a function of $a$ for $\ell_1$- and $\ell_2$-ARD. The plain line is the average value of $K_\text{eff}$ over the 10 runs and dashed-lines display $\pm$ the standard deviation.}
   \label{fig:swim_paths}
   \end{center}
\end{figure}

\begin{figure}[t]
 \begin{center}
   \includegraphics[width=\linewidth]{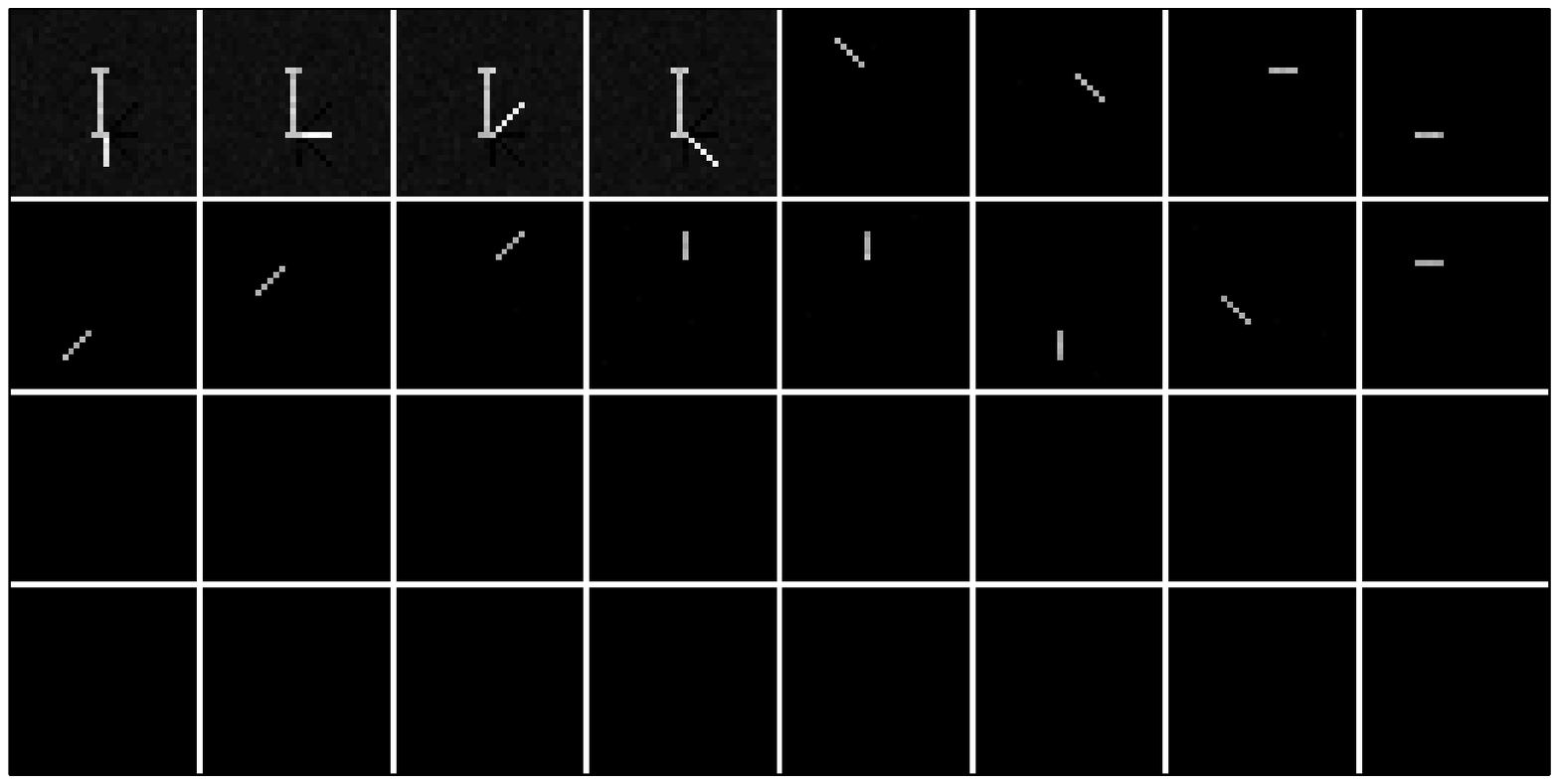}\\
   \includegraphics[width=\linewidth]{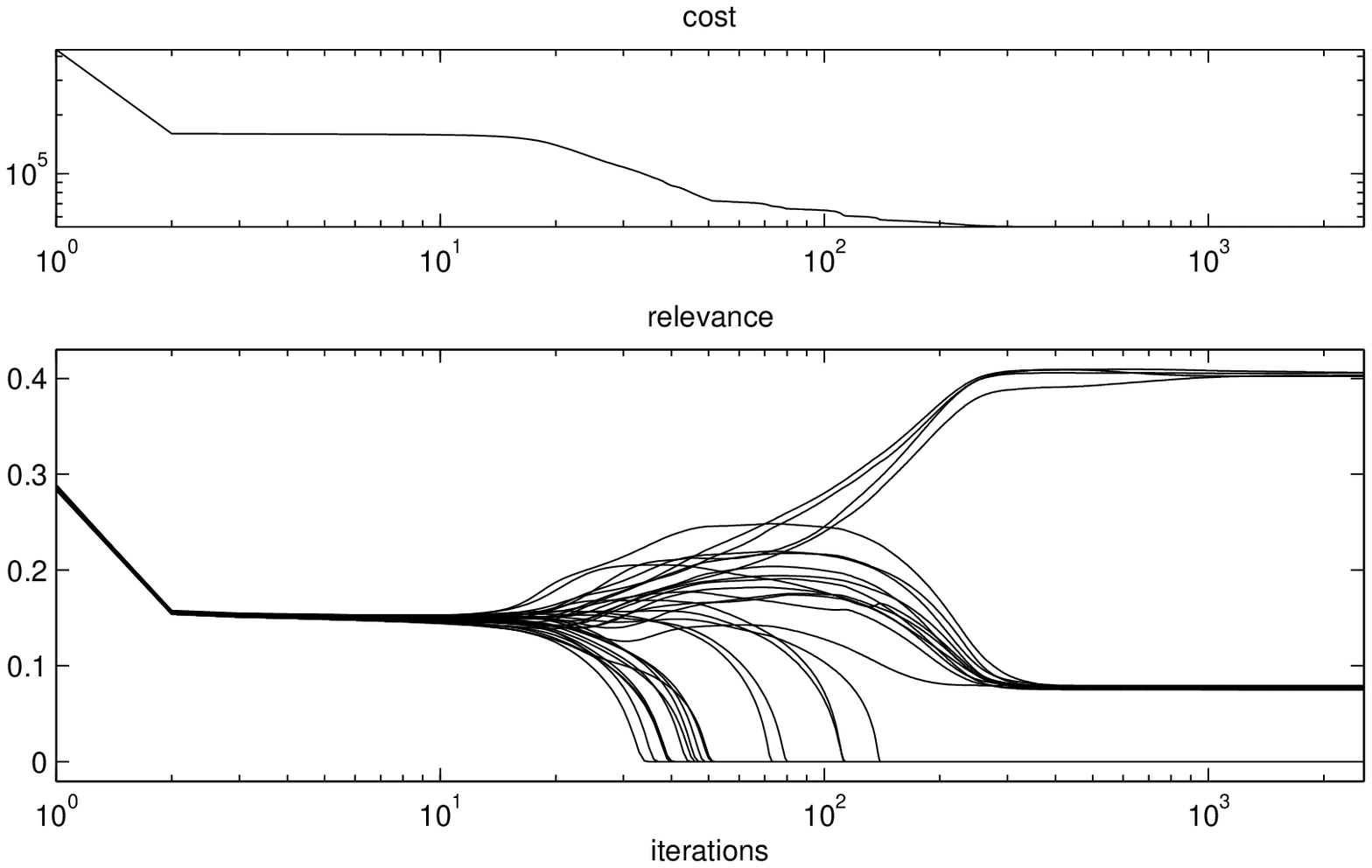}   
   \caption{Top: Dictionary learnt in one run of $\ell_1$-ARD with $a=100$. The dictionary elements are presented left to right, top to bottom, by descending order of their relevance $\lambda_k$. For improved visualization and fair comparison of the relative importance of the dictionary elements, we display $\bw_k$ rescaled by the expectation of $h_{kn}$, i.e., for $\ell_1$-ARD, $\lambda_k \bw_k$. The figure colormap is then adjusted to fit the full range of values taken by $\ve{W} \diag{\boldsymbol{\lambda}}$. Middle: values of the objective function~\eqref{eqn:lasso_cost} along iterations (log-log scale). Bottom: values of $\lambda_k - B$ along iterations (log-linear scale).}
   \label{fig:swim_l1}
   \end{center}
\end{figure}

We applied $\ell_1$- and $\ell_2$-ARD with $\beta=1$ (KL-divergence, matching the Poisson noise assumption, and thus $\phi =1$), $K = 32 = 2\, K_{\mathrm{true}}$ and $\tau = 10^{-6}$. We tried several values for the hyperparameter $a$, namely $a \in \{ 5, 10, 25, 50, 75, 100, 250, 500, 750, 1000 \}$, and set $b$ according to \eqref{eqn:choose_b}. For every value of $a$ we ran the algorithms   from 10 common positive random initializations. The regularization paths returned by the two algorithms are displayed in Fig.~\ref{fig:swim_paths}. $\ell_1$-ARD consistently estimates the correct number of components ($K_{\mathrm{true}}=16$) up to $a = 500$. Fig.~\ref{fig:swim_l1} displays the learnt basis, objective function and relevance parameters along iterations in one run of $\ell_1$-ARD when $a=100$. It can be seen that the ground truth is perfectly recovered. 

\begin{figure}[t]
 \begin{center}
   \includegraphics[width=\linewidth]{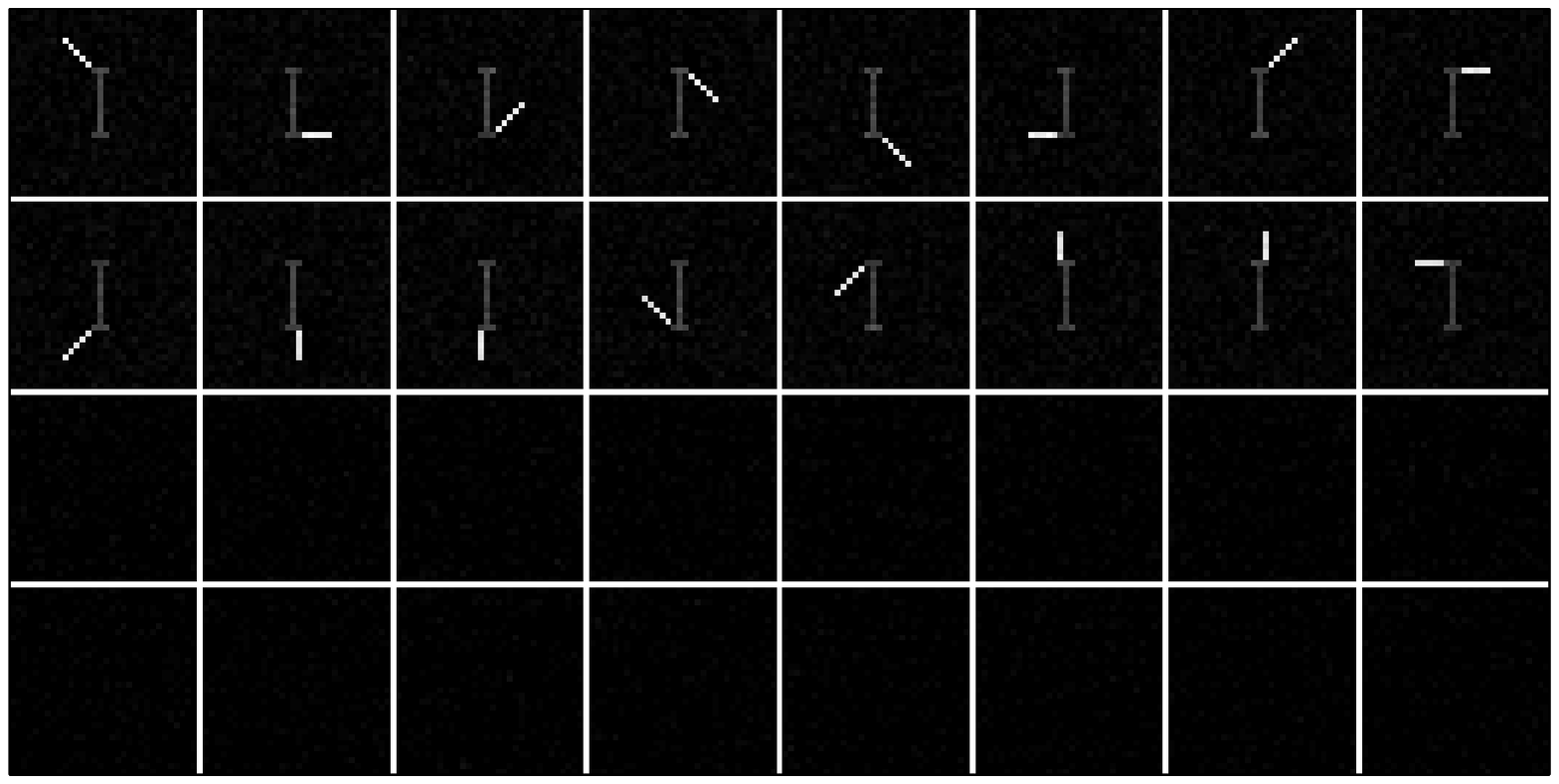}\\
   \includegraphics[width=\linewidth]{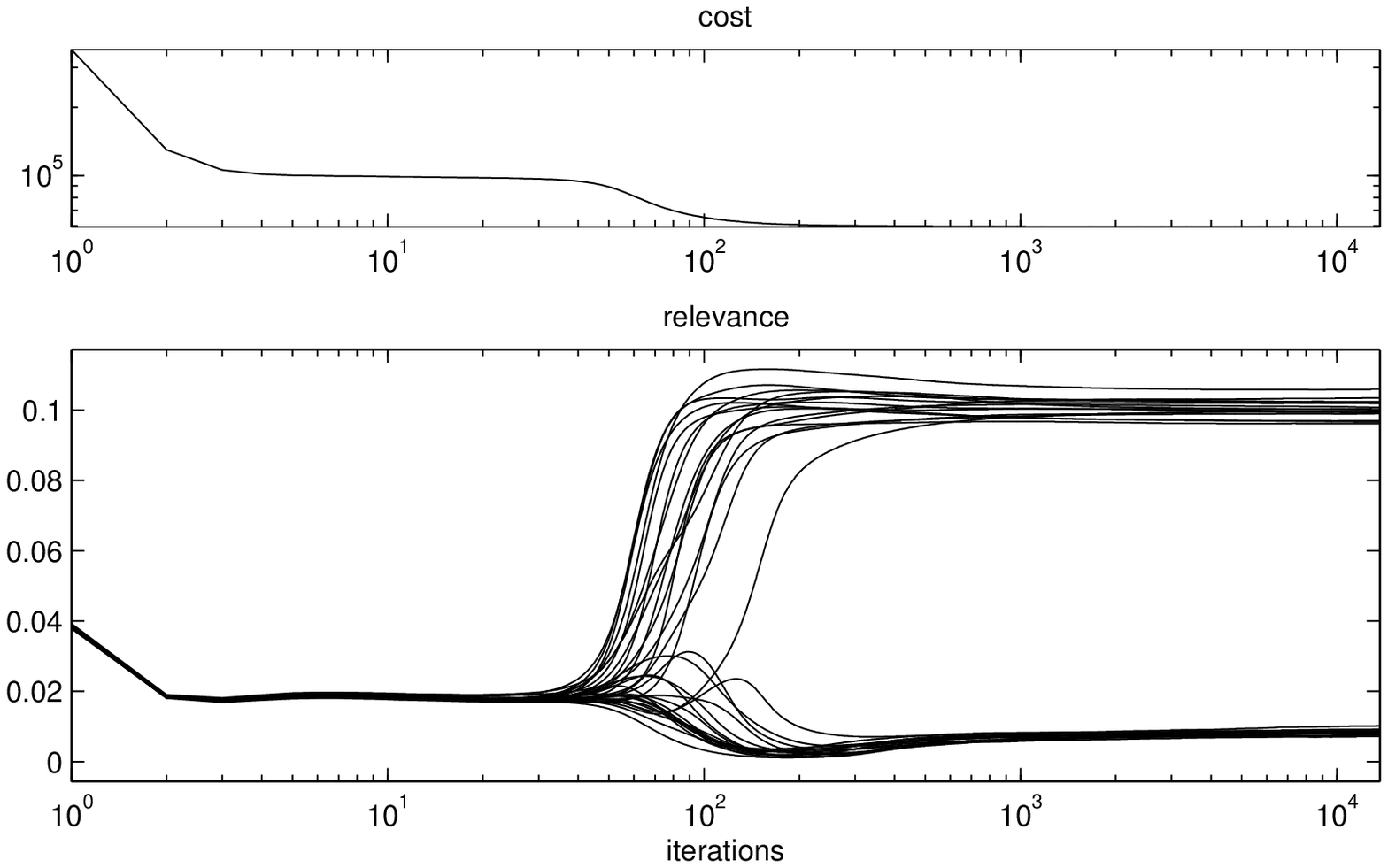}       
   \caption{Top: Dictionary learnt by $\ell_2$-ARD with $a=1000$. The dictionary is displayed using the same convention as in Fig.~\ref{fig:swim_l1}, except that the vectors $\bw_k$ are now rescaled by the expectation of $h_{kn}$ under the Half-Normal prior, i.e., $(2\lambda_k/\pi)^{1/2}$. Middle: values of the cost function~\eqref{eqn:lasso_cost} along iterations (log-log scale). Bottom: values of $\lambda_k - B$ along iterations (log-linear scale).}
   \label{fig:swim_l2}
   \end{center}
\end{figure}

In contrast to $\ell_1$-ARD, Fig.~\ref{fig:swim_paths} shows that the value of $K_\text{eff}$ returned by $\ell_2$-ARD is more variable across runs and values of $a$. Manual inspection reveals that some runs return the correct decomposition when $a=500$ (and those are the runs with lowest end value of the objective function, indicating the presence of apparent local minima), but far less consistently than $\ell_1$-ARD. Then it might appear like the decomposition strongly overfits the noise for $a \in \{750, 1000\}$. However, visual inspection of learnt dictionaries with these values show that the solutions still make sense. As such, Fig.~\ref{fig:swim_l2} displays the dictionary learnt by $\ell_2$-ARD with $a=1000$. The figure shows that the hierarchy of the decomposition is preserved, despite that the last 16 components capture some residual noise, as a closer inspection would reveal. Thus, despite that pruning is not fully achieved in the 16 extra components, the relevance parameters still give a valid interpretation of what are the most significant components. Fig.~\ref{fig:swim_l2} shows the evolution of relevance parameters along iterations and it can be seen that the 16 ``spurious'' components   approach the lower bound in the early iterations before they start to fit  noise. Note that $\ell_2$-ARD returns a solution where the torso is equally shared by the four limbs. This is because the $\ell_2$ penalization  favors this particular solution over the one returned by $\ell_1$-ARD, which favors sparsity of the individual dictionary elements.

With $\tau = 10^{-6}$, the average number of iterations for convergence is  approximately $ 4000\pm2000$ for $\ell_1$-ARD   for all $a$.  The average number of iterations for $\ell_2$-ARD is of the same order for $a \le 500$, and increases to more than $10,000$ iterations for $a\ge 750$, because all components are active for these $a$'s.

\subsection{Music decomposition } \label{sec:audio}

\begin{figure}[t]
 \begin{center}
   \includegraphics[width=\linewidth]{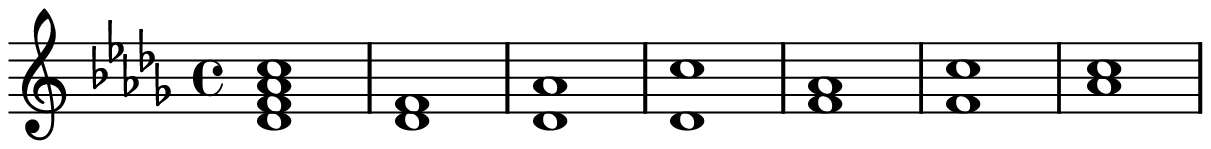}    \\
     \includegraphics[width=\linewidth]{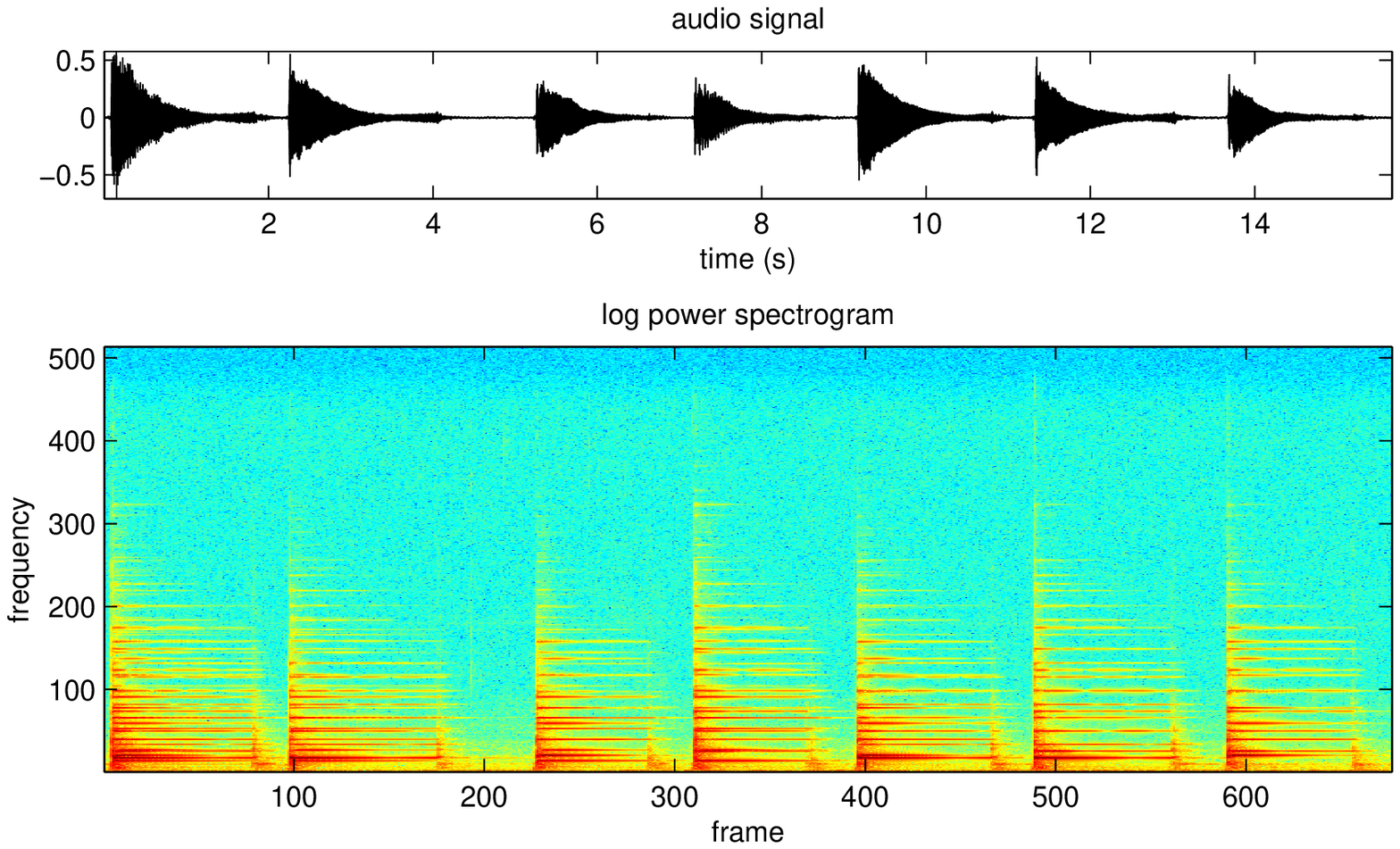}
   \end{center}     
   
   \caption{Three representations of data; (top): original score, (middle): time-domain recorded signal, (bottom): log-power spectrogram.}
   \label{fig:piano_data}
\end{figure}

We now consider a music signal decomposition example and illustrate the benefits of ARD in NMF with the IS divergence ($\beta = 0$). F\'{e}votte et al.~\cite{neco09} showed that IS-NMF of the power spectrogram underlies a generative statistical model of superimposed Gaussian components, which is   relevant to the representation of audio signals. As explained in Sections~\ref{sec:likel} and~\ref{sec:selectab}, this model is also equivalent to assuming that the power spectrogram is observed in multiplicative exponential noise, i.e., setting $\phi\! =\! 1/\alpha \!=\! 1$. We investigate the decomposition of the short piano sequence used in \cite{neco09}, a monophonic 15 seconds-long signal $x_t$ recorded in real conditions. The sequence is composed of 4 piano notes, played all at once in the first measure and then played by pairs in all possible combinations in the subsequent measures. The STFT $x_{fn}$ of the temporal data $x_t$ was computed using a sinebell analysis window of length $L = 1024$ (46 ms) with 50 \% overlap between two adjacent frames, leading to $N = 674$ frames and $F = 513$ frequency bins. The musical score, temporal signal and log-power spectrogram are shown in Fig.~\ref{fig:piano_data}. In \cite{neco09} it was shown that IS-NMF of the power spectrogram $v_{fn} = |x_{fn} |^2$ can correctly separate the spectra of the different notes and  other constituents of the signal (sound of hammer on the strings, sound of sustain pedal etc.).

We set $K = 18$ (3 times the ``ground truth" number of components) and ran $\ell_2$-ARD with $\beta =0$, $a = 5$ and $b$ computed according to~\eqref{eqn:choose_b}. We ran the algorithm from 10 random initializations and selected the solution returned with the lowest final cost. For comparison, we ran standard nonpenalized Itakura-Saito NMF using the multiplicative algorithm described in \cite{neco09}, equivalent to $\ell_2$-ARD  with $\lambda_k \rightarrow \infty$ and $\gamma(\beta) =1$. We ran IS-NMF 10 times with the same random initializations we used for ARD IS-NMF, and selected the solution with minimum fit. Additionally, we ran the methods by M\o rup \& Hansen (with KL-divergence) \cite{mor09} and Hoffman \textit{et al.} \cite{hoffman}. We used Matlab implementations either publicly available \cite{hoffman}  or provided to us by the authors \cite{mor09}. The best among ten runs of these methods was selected. 

Given an approximate factorization $\ve{W} \ve{H}$ of the data spectrogram $\ve{V}$ returned by any of the four algorithms we proceeded to reconstruct time-domain components by Wiener filtering, following \cite{neco09}. The STFT estimate $\hat{c}_{k,fn}$ of component $k$ is reconstructed by
\bal{
\hat{c}_{k,fn} = \frac{w_{fk} h_{kn}}{\sum_j w_{fj} h_{jn}} x_{fn}
}
and the STFT is inverted to produce the temporal component $\hat{c}_{k,t}$.\footnote{With the approach of Hoffman \textit{et al.} \cite{hoffman}, the columns of $\ve{W}$ have to be multiplied by their corresponding weight parameter $\theta_k$ prior to reconstruction.} By linearity of the reconstruction and inversion, the decomposition is conservative, i.e., 
$
x_t \!=\! \sum_k \hat{c}_{k,t}.$

\begin{figure}
 \begin{center}
 \begin{minipage}{0.49\linewidth}
 \begin{center}
 {\small (a) IS-NMF}
 
 \smallskip
    \includegraphics[width=0.99\linewidth]{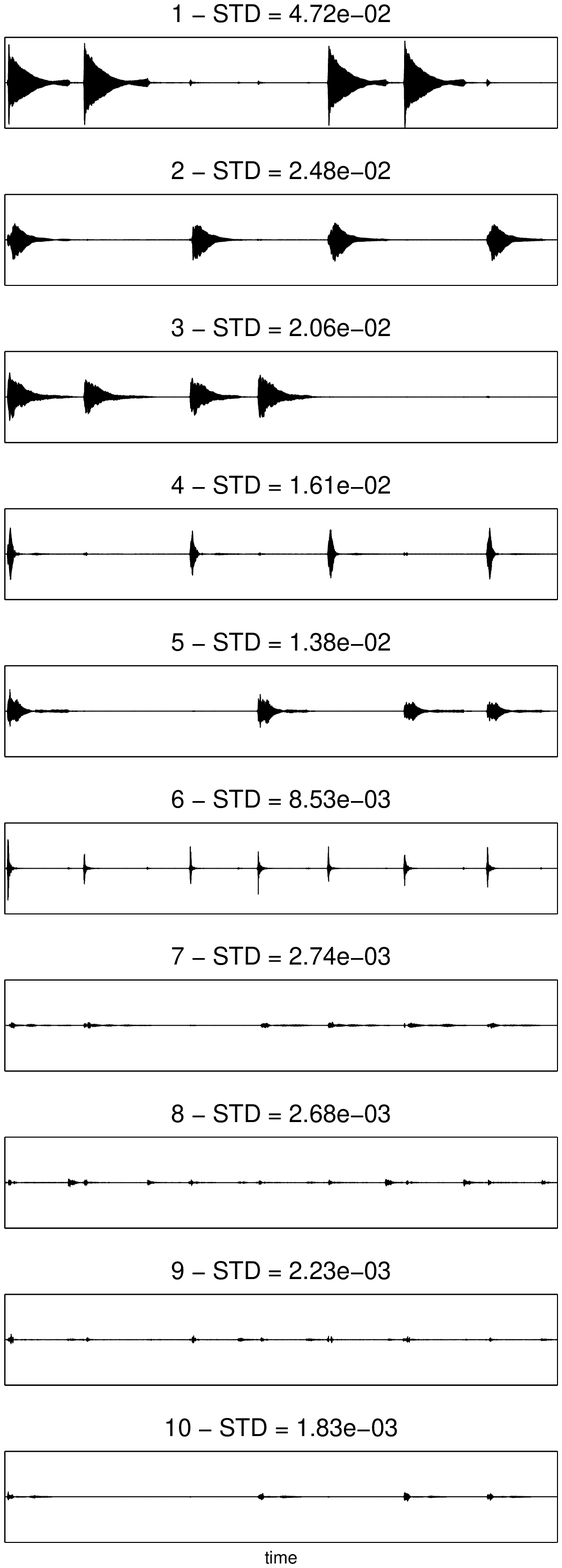}    
    \end{center}
   \end{minipage}
 \begin{minipage}{0.49\linewidth}
  \begin{center}
{\small (b) ARD IS-NMF}

 \smallskip
   \includegraphics[width=0.99\linewidth]{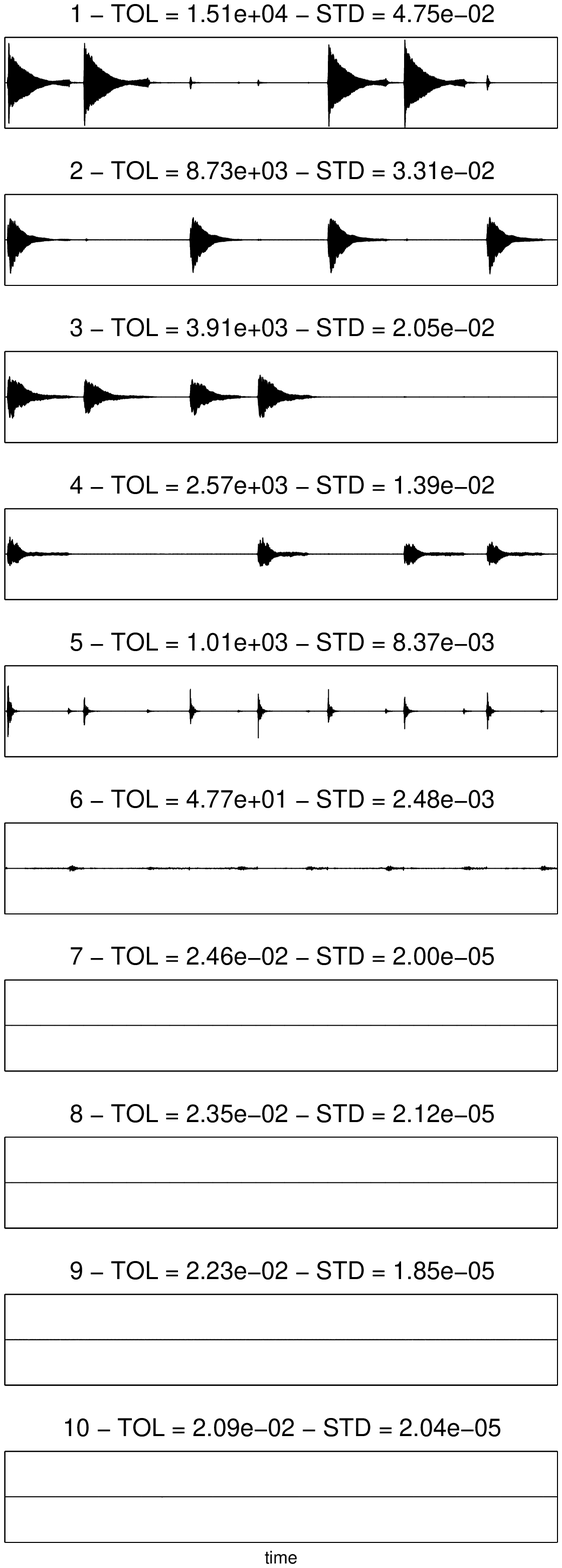}    
   \end{center}
   \end{minipage}   
   \caption{The first ten components produced by IS-NMF and ARD IS-NMF. STD denotes the standard deviation of the   time samples. TOL is the relevance relative to the bound, i.e., $(\lambda_k - B)/B$. With IS-NMF, the second note of the piece is split into two components ($k=2$ and $k=4$).}
   \label{fig:piano_comps}
   \end{center}
\end{figure}

\begin{figure}
 \begin{center}
   \includegraphics[width=\linewidth]{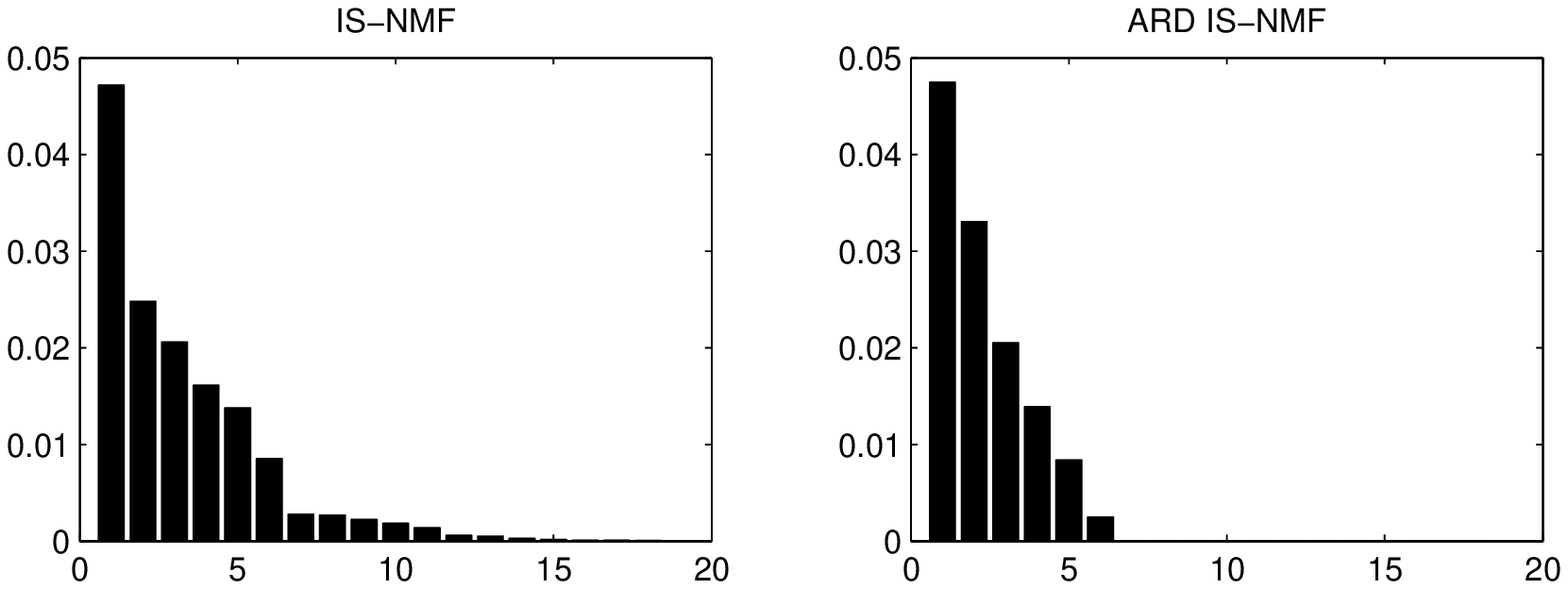}  \\
      \includegraphics[width=\linewidth]{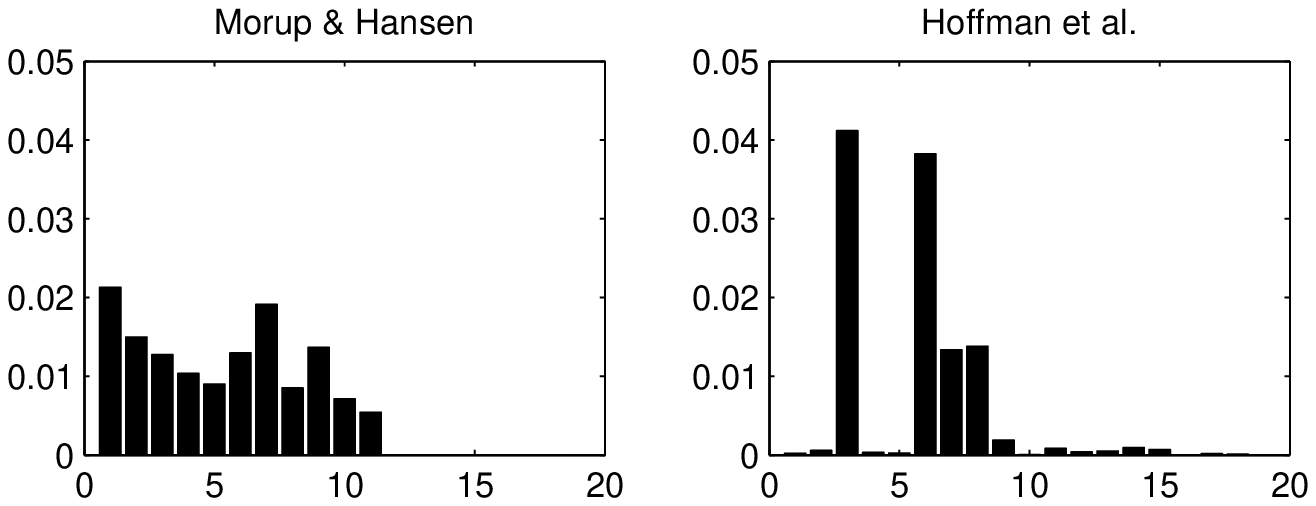}    
   \caption{Histograms of standard deviation values of all 18 components produced by IS-NMF, ARD IS-NMF, M\o rup \& Hansen \cite{mor09} and Hoffman {\it et al.} \cite{hoffman}. ARD IS-NMF only retains 6 components, which correspond to the expected decomposition, displayed in Fig.~\ref{fig:piano_comps}. On this dataset, the methods proposed in \cite{mor09} and \cite{hoffman} fail to produce the desired decomposition.}
   \label{fig:piano_hists}
   \end{center}
\end{figure}

The components produced by IS-NMF were ordered \emph{by decreasing value of their standard deviations} (computed from the time samples). The components produced by ARD IS-NMF, M\o rup \& Hansen \cite{mor09} and Hoffman {\it et al.} \cite{hoffman} were ordered \emph{by decreasing value of their relevance weights} ($\{\lambda_k\}$ or $\{\theta_k \}$). Fig.~\ref{fig:piano_comps} displays the ten first components produced by IS-NMF and ARD IS-NMF. The $y$-axes of the two figures are identical so that the component amplitudes are directly comparable. Fig.~\ref{fig:piano_hists} displays the histograms of the standard deviation values of all 18 components for IS-NMF, ARD IS-NMF, M\o rup \& Hansen \cite{mor09} and Hoffman {\it et al.} \cite{hoffman}.\footnote{The sound files produced by all the approaches are available at \url{http://perso.telecom-paristech.fr/~fevotte/Samples/pami12/soundsamples.zip}}

The histogram on top right of Fig.~\ref{fig:piano_hists} indicates that ARD IS-NMF retains   6 components. This is also confirmed by the value of relative relevance $(\lambda_k - B)/B$ (upon convergence of the relevance weights), displayed with the components on Fig.~\ref{fig:piano_comps}, which drops by a factor of about 2000 from component 6 to component 7. The 6 components correspond to expected semantic units of the musical sequence: the first four components extract the individual notes and the next two components extract the sound of hammer hitting the strings and the sound produced by the sustain pedal when it is released. In contrast IS-NMF has a tendency to overfit, in particular the second note of the piece is split into two components ($k=2$ and $k=4$). The histogram on bottom left of Fig.~\ref{fig:piano_hists} shows that the approach of M\o rup \& Hansen \cite{mor09} (with the KL-divergence) retains 11 components. Visual inspection of the reconstructed components reveals inaccuracies in the decomposition and significant overfit (some notes are split in subcomponents). The poorness of the results is in part explained by the inadequacy of the KL-divergence (or Euclidean distance) for factorization of spectrograms, as discussed  in~\cite{neco09}. In contrast our approach offers flexibility for ARD NMF where the fit-to-data term can be chosen according to the application by   setting $\beta$ to the desired value.

The histogram on bottom right of Fig.~\ref{fig:piano_hists} shows that  the method by Hoffman {\it et al.} \cite{hoffman}  retains approximately 5 components. The decomposition resembles the expected decomposition more closely than \cite{mor09}, except that the hammer attacks  are merged with one of the notes. However it is interesting to note that the distribution of standard deviations does not follow the order of relevance values. This is because the weight parameter $\theta_k$ is independent of $\ve{W}$ and $\ve{H}$ in the prior. As such, the factors are allowed to take very small values while the weight values are not necessarily small. 

Finally, we remark that on this data $\ell_1$-ARD IS-NMF performed similarly to $\ell_2$-ARD IS-NMF and in both cases the retrieved decompositions were  fairly robust to the choice of $a$. We experimented with the same values of $a$ as in previous section and the decompositions and their hierarchies were always found correct. We point out that, as with IS-NMF, initialization is an issue, as other runs did not produced the desired decomposition into notes. However, in our experience the best out of 10 runs   always output the correct decomposition.

\subsection{Prediction of stock prices} \label{eqn:stock}

\begin{figure}
 \begin{center}
\includegraphics[width=\linewidth]{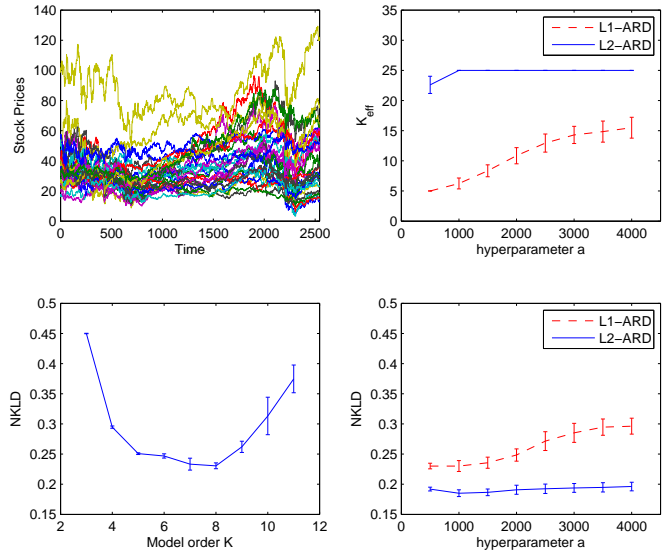}           
   \caption{Top left: The stock data; Top right: Effective model order $\Keff$ as a function of $a$. Bottom:   Normalized KL-divergence (NKLD) for KL-NMF (left), $\ell_1$- and $\ell_2$-ARD KL-NMF (right). Note that the $y$-axes on both plots are the same. M\o rup \& Hansen's method~\cite{mor09}  yielded an NKLD of $0.37 \pm 0.03$ (averaged over 10 runs) which is inferior to $\ell_2$-ARD in as seen on the bottom right.}
   \label{fig:dow-kl}
   \end{center}
\end{figure} 

NMF (with the Euclidean and KL costs) has previously been applied on stock data~\cite{dra07} to learn ``basis functions'' and to cluster companies. In this section, we perform a prediction task on the stock prices of the Dow 30 companies (comprising the Dow Jones Industrial Average). These  are  major American companies   from various sectors of the economy such as services (e.g., Walmart), consumer goods (e.g., General Motors) and healthcare (e.g., Pfizer). The dataset consists of the stock prices of these $F=30$ companies from 3rd Jan 2000 to 27th Jul 2011, a total of $N= 2543$ trading days.\footnote{Stock prices of the Dow 30 companies are provided at the following link: \url{http://www.optiontradingtips.com/resources/historical-data/dow-jones30.html}. The raw data consists of 4 stock prices per company per day. The mean of the 4    data points is taken to be the representative of the stock price of that company for that day.}  The data are displayed in the top left plot of Fig.~\ref{fig:dow-kl}. 

In order to test the prediction capabilities of our algorithm, we organized the data into an $F\times N$ matrix $\bV$ and removed $50\%$ of the entries at random.  For the first set of experiments, we performed standard $\beta$-NMF with $\beta=1$, for different values of $K$, using the observed entries only.\footnote{Accounting for the missing data involves applying a binary mask to $\ve{V}$ and $\ve{W} \ve{H}$, where 0 indicates missing entries~\cite{ho08}.} We report results for different non-integer  values of $\beta$ in the following. Having performed KL-NMF on the incomplete data, we then estimated the missing entries by multiplying the inferred basis $\bW$ and the activation coefficients $\bH$ to obtain the estimate $\hat{\bV}$. The normalized KL-divergence (NKLD) between the   true (missing) stock data and their estimates is then computed as
\begin{equation}
\mbox{NKLD}\defeq \frac{1}{|\calE|} \sum_{(f,n)\in \calE} d_{\mathrm{KL}}(v_{fn}| \hatv_{fn}), \label{eqn:def_nkld}
\end{equation}
where $\calE \subset \{1,\ldots, F\}\times \{1,\ldots, N\}$ is the set of missing entries and $d_{\mathrm{KL}}(\fndot|\fndot)$ is the KL-divergence ($\beta=1$). The smaller the NKLD, the better the prediction of the missing stock prices and hence the better the decomposition of $\bV$ into $\bW$ and $\bH$. We then did the same for $\ell_1$- and $\ell_2$-ARD KL-NMF, for different values of  $a$ and using $K=25$. For  KL-NMF   the criterion for termination is chosen so that it mimics that in Section~\ref{sec:stopping}. Namely, as is commonly done in the NMF literature, we ensured that the columns of $\bW$ are normalized to unity. Then, we computed the  {\em NMF relevance weights} $\lambda_k^{\mathrm{NMF}} \defeq\frac{1}{2} \|\underline{h}_k\|_2^2$. We terminate the algorithm whenever
$
\mathrm{tol}^{\mathrm{NMF}}\defeq\max_{k} |{ (\lambda_k^{\mathrm{NMF}} - \tlambda_k^{\mathrm{NMF}} ) }/{ \tlambda_k^{\mathrm{NMF}} } |
$
falls below   $\tau = 5\times 10^{-7}$. We averaged the results over 20 random initializations.  The NKLDs  and the the inferred   model orders $\Keff$     are   displayed in   Fig.~\ref{fig:dow-kl}.

In the top right plot of Fig.~\ref{fig:dow-kl}, we observe that there is a general   increasing trend; as $a$   increases, the inferred model order $\Keff$ also increases. In addition, for the same value of $a$, $\ell_1$-ARD prunes more components than $\ell_2$-ARD due to its   sparsifying effect. This was also observed for synthetic data and the \texttt{swimmer} dataset. However, even though $\ell_2$-ARD retains  almost all the components, the basis and activation coefficients learned model the underlying data better. This is because $\ell_2$ penalization methods result  in coefficients that are more dense and are known to be better for prediction (rather than sparsification) tasks.

From the bottom left plot of Fig.~\ref{fig:dow-kl}, we observe that  when $K$ is too small, the model is not ``rich'' enough to model the data and hence the NKLD is large. Conversely, when $K$ is too large, the model overfits the data, resulting in a large NKLD.  We also observe that $\ell_2$-ARD performs spectacularly across a range of values of the hyperparameter $a$, uniformly better than standard KL-NMF. The NKLD for estimating the missing stock prices hovers around 0.2, whereas KL-NMF results in an NKLD of more than 0.23   for all $K$.  This shows that $\ell_2$-ARD produces a decomposition that is more relevant for modeling   missing data. Thus, if one does not know the true model order {\em a priori} and chooses  to use  $\ell_2$-ARD with some hyperparameter $a$, the resulting NKLD would be much better than doing KL-NMF even though many components will be retained.  In contrast, $\ell_1$-ARD does not perform as spectacularly across all values of $a$ but even when a small number of components is retained (at $a=500$, $\Keff=5$, NKLD for $\ell_1$-ARD $\approx 0.23$, NKLD for KL-NMF $\approx 0.25$), it performs significantly better than KL-NMF. It is plausible that the  stock data fits  the assumptions of the Half-Normal model better than the Exponential  model and hence $\ell_2$-ARD performs better. 

For comparison, we also implemented a version of the method by M\o rup \& Hansen~\cite{mor09} that handles missing data. The mean NKLD value returned  over ten runs is $0.37 \pm 0.03$, and thus  it is   clearly inferior to the   methods in this paper.   The data does not fit the   model well.

\begin{figure}
 \begin{center}        
\includegraphics[width=\linewidth]{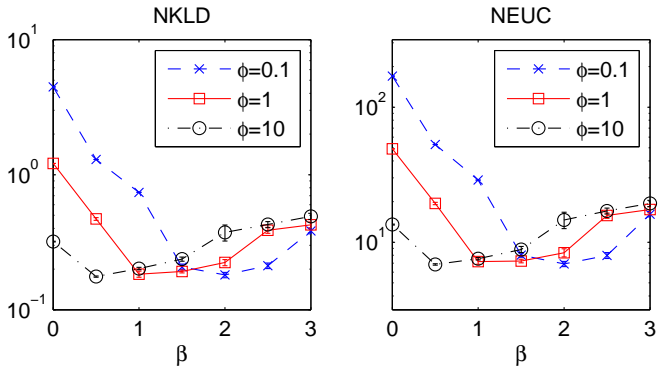}           
   \caption{Effect of varying shape $\beta$ and dispersion $\phi$ on prediction performance. Average results over 10 runs.}  
   \label{fig:vary_beta}
   \end{center}
\end{figure} 

Finally, in Fig.~\ref{fig:vary_beta} we demonstrate the effect of varying the shape parameter  $\beta$   and the dispersion   parameter $\phi$. The distance  between  the predicted stock prices and the   true ones is measured using the NKLD in~\eqref{eqn:def_nkld} and the NEUC (the Euclidean analogue of the NKLD). We also computed the NIS (the IS analogue of the NKLD), and noted that the results across all $3$ performance metrics are   similar so we omit the  NIS. We used $l_2$-ARD, set $a=1000$ and calculated $b$ using~\eqref{eqn:choose_b}. We also chose integer and non-integer values of $\beta$ to demonstrate the flexibility of $l_2$-ARD. It is observed that $\beta=0.5,\phi=10$ gives the best NKLD and NEUC and that $1\le \beta\le  1.5$ performs well across a wide range of values of $\phi$.

\section{Conclusion} \label{sec:concl}
In this paper, we proposed a  novel statistical model for $\beta$-NMF where the columns of $\ve{W}$ and rows $\ve{H}$ are tied together through a common scale parameter in their prior, exploiting (and solving) the scale ambiguity between $\ve{W}$ and $\ve{H}$. MAP estimation reduces to a penalized NMF problem with a   group-sparsity inducing regularizing term. A set of MM algorithms accounting for all values of $\beta$ and either $\ell_1$- or $\ell_2$-norm group-regularization was presented. They ensure the monotonic decrease of the objective function at each iteration and result in multiplicative update rules of linear complexity in $F, K$ and $N$. The updates automatically preserve nonnegativity given positive initializations and are easily implemented. The efficiency of our approach was validated on several synthetic and real-world datasets, with competitive performance w.r.t.\ the  state-of-the-art. At the same time, our proposed methods offer  improved flexibility over existing approaches (our approach can deal with various types of observation noise and prior structure in a unified framework). Using the method of moments, an effective strategy for the selection of hyperparemeter $b$ given $a$ was proposed and, as a general rule of thumb, we recommend to set $a$ a small value w.r.t.\ $F+N$.

There are several avenues for further  research: Here we derived a MAP approach that works efficiently, but more sophisticated inference techniques can be envisaged, such as fully Bayesian inference in the  model we proposed in Section~\ref{sec:ardnmf}. Following similar treatments in sparse regression~\cite{tip01,Wipf} or with other forms of matrix factorization \cite{Sal07}, one could seek the maximization of $\log p(\ve{V} | a, b, \phi)$ using variational or Markov chain Monte-Carlo inference, and in particular handle hyperparameter estimation in a (more) principled way. Other more direct extensions of this work  concern the factorization of tensors and online-based methods akin to~\cite{Lefevre,mair10}.

\subsubsection*{Acknowledgements} The authors would like to acknowledge Francis Bach for discussions related to this work, Y.~Kenan~Y\i lmaz and A.~Taylan Cemgil for discussions on Tweedie  distributions, as well as Morten M\o rup and Matt Hoffman for sharing their code. We would also like to thank the   reviewers whose comments helped to greatly improve the paper.

\vspace{-5mm}
\begin{IEEEbiography}[{\includegraphics[width=1in,height
=1.25in,clip,keepaspectratio]{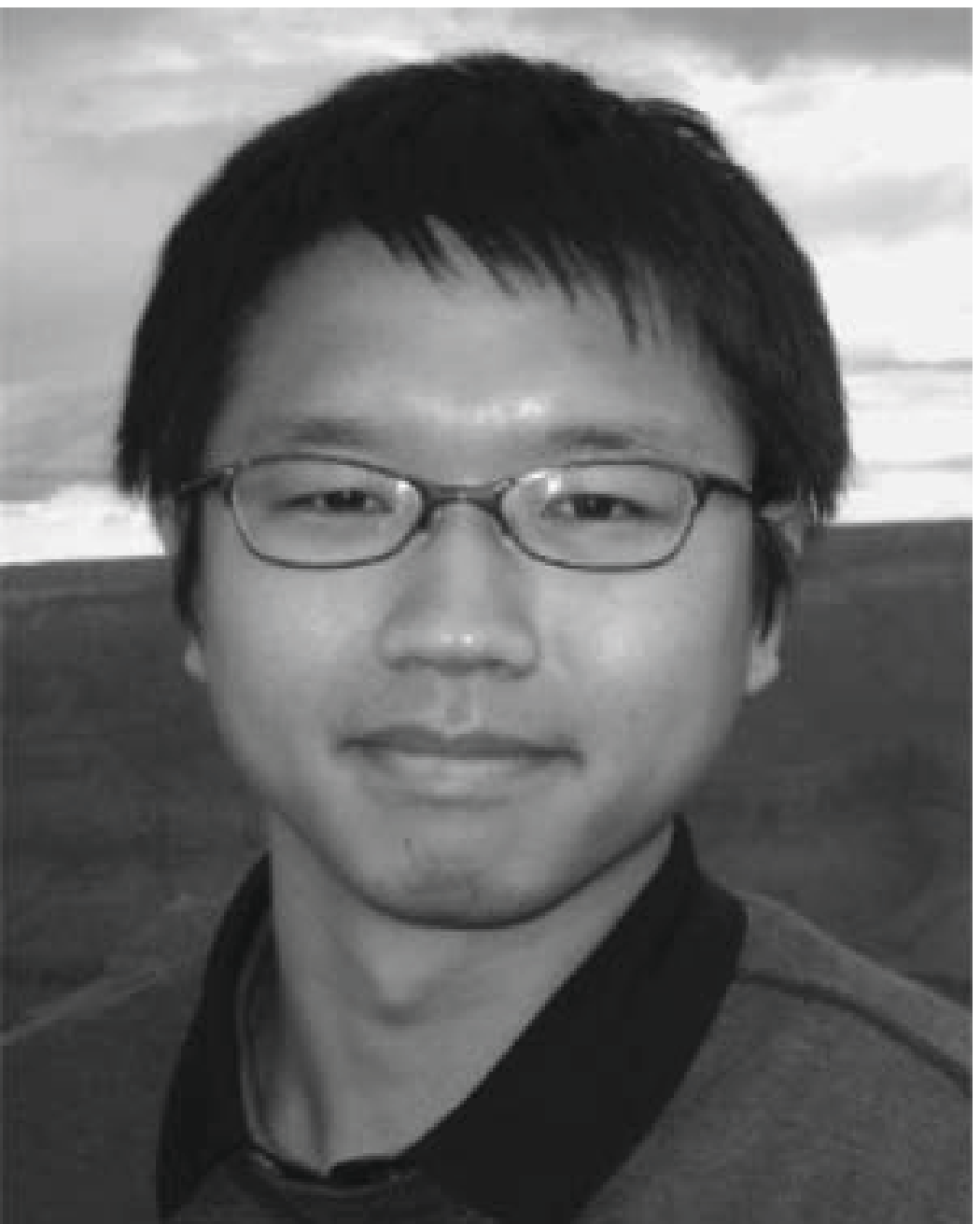}}]{\bf
Vincent Y. F. Tan}    did the Electrical and Information Sciences Tripos (EIST) at the University of Cambridge and obtained a B.A.\ and an  M.Eng.\ degree  in 2005. He received a Ph.D.\ degree in Electrical Engineering and Computer Science from MIT in 2011. After which, he was a postdoctoral researcher at University of Wisconsin-Madison. He is now a scientist at the Institute for Infocomm Research (I$^2$R), Singapore and an adjunct assistant professor at the department of Electrical and Computer Engineering at the National University of Singapore. He held two summer research internships at Microsoft Research during his Ph.D.\ studies. His  research interests include learning and inference in graphical models, statistical signal processing and   network information theory. Vincent received the Charles Lamb prize, a Cambridge University Engineering Department prize awarded to the student who demonstrates the greatest proficiency in the EIST. He also received the MIT EECS Jin-Au Kong outstanding doctoral thesis prize  and the A*STAR Philip Yeo prize for outstanding achievements in research.
\end{IEEEbiography}
\vspace{-5mm}
\begin{IEEEbiography}[{\includegraphics[width=1in,height
=1.25in,clip,keepaspectratio]{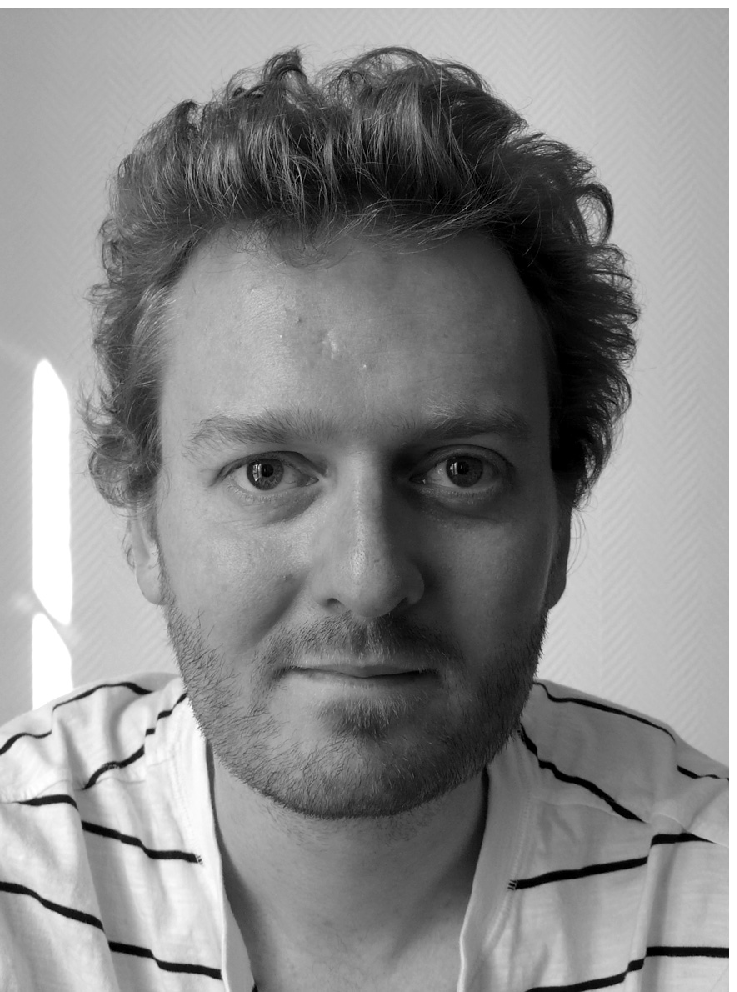}}]
{C\'edric F\'evotte} obtained the State Engineering degree and the PhD degree in Control and Computer Science from \'Ecole Centrale de Nantes (France) in 2000 and 2003, respectively. As a PhD student he was with the Signal Processing Group at Institut de Recherche en Communication et Cybern\'etique de Nantes (IRCCyN). From 2003 to 2006 he was a research associate with the Signal Processing Laboratory at University of Cambridge (Engineering Dept). He was then a research engineer with the music editing technology start-up company Mist-Technologies (now Audionamix) in Paris. In Mar. 2007, he joined T\'el\'ecom ParisTech, first as a research associate and then as a CNRS tenured research scientist in Nov. 2007. His research interests generally concern statistical signal processing and unsupervised machine learning and in particular applications to blind source separation and audio signal processing. He is the scientific leader of project TANGERINE (Theory and applications of nonnegative matrix factorization) funded by the French research funding agency ANR and a member of the IEEE ``Machine learning for signal processing" technical committee.
\end{IEEEbiography}

\end{document}